\pdfoutput=1

\documentclass[11pt]{article}

\usepackage[preprint]{acl}

\usepackage{times}
\usepackage{latexsym}

\usepackage[T1]{fontenc}

\usepackage[utf8]{inputenc}

\usepackage{microtype}

\usepackage{inconsolata}

\usepackage{graphicx}

\usepackage{tabularx}
\usepackage[normalem]{ulem}
\usepackage{amssymb}
\usepackage{amsfonts}
\usepackage{multicol}
\usepackage{tipa}
\usepackage{bm}
\usepackage[ruled,linesnumbered]{algorithm2e}
\usepackage{paralist}
\usepackage{IEEEtrantools}
\usepackage[switch]{lineno}
\usepackage{enumitem}
\usepackage{arydshln}
\usepackage{verbatim}
\usepackage{pifont}
\usepackage{csquotes}
\usepackage{xspace}
\usepackage{mdwlist}
\usepackage{subfigure}
\usepackage{array}
\usepackage{colortbl}
\usepackage{dsfont}
\usepackage{url}
\usepackage{booktabs} 
\usepackage{tabularx}
\usepackage{bbm}
\usepackage{pgfplots}
\usepackage{tikz}
\usepackage{soul} 
\usepackage{color,xcolor} 
\usepackage{listings}
\usepackage{multirow}
\usepackage{amsmath}
\usepackage{longtable}
\usepackage{supertabular}
\usepackage{makecell}
\usepackage{subfloat}

\newcommand{\red}[1]{\textcolor{red}{#1}}

\title{Haste Makes Waste: Evaluating Planning Abilities of LLMs for \\ Efficient and Feasible  Multitasking with Time Constraints Between Actions}

\author{
    Zirui Wu$^{1}$, Xiao Liu$^{1}$, Jiayi Li$^{1}$,
    Lingpeng Kong$^{2}$\and
    Yansong Feng$^{1}$\thanks{\;\;Corresponding authors.}\\
    $^1$Peking University, 
    $^2$The University of Hong Kong\\
    {\tt ziruiwu@pku.edu.cn}, 
    {\tt fengyansong@pku.edu.cn}\\
}

\begin{document}
\maketitle
\begin{abstract}
While Large Language Model-based agents have demonstrated substantial progress in task completion, existing evaluation benchmarks tend to overemphasize single-task performance, with insufficient attention given to the crucial aspects of multitask planning and execution efficiency required in real-world scenarios. To bridge this gap, we present \textsc{Recipe2Plan}, a novel benchmark framework based on real-world cooking scenarios. Unlike conventional benchmarks, \textsc{Recipe2Plan} challenges agents to optimize cooking time through parallel task execution while respecting temporal constraints i.e. specific actions need to be performed within a particular time
intervals following the preceding steps.
Overly aggressive local parallelization may disrupt this constraint, potentially compromising the entire cooking process.
This strict time constraint between actions raises a unique challenge for agents 
to balance between maximizing concurrent operations and adhering to critical timing constraints. Extensive experiments with state-of-the-art models reveal challenges in maintaining this balance between efficiency and feasibility. The results highlight the need for improved temporal awareness and global multitasking capabilities in large language models. We open-source our benchmark and code at \url{https://github.com/WilliamZR/Recipe2Plan}.
\end{abstract}

\section{Introduction}

Large Language Models (LLMs)~\citep{openai2023gpt4,geminiteam2023gemini,touvron2023llama,qwen2.5} have demonstrated the ability to plan and reason step by step~\citep{wei2022chain}.  Leveraging this ability, LLM-based agents can automate complex real-world tasks~\cite{yao2022react, shinn2024reflexion, sun2024adaplanner}. 


\begin{figure*}[t]
    \centering
    \includegraphics[width=0.85\linewidth]{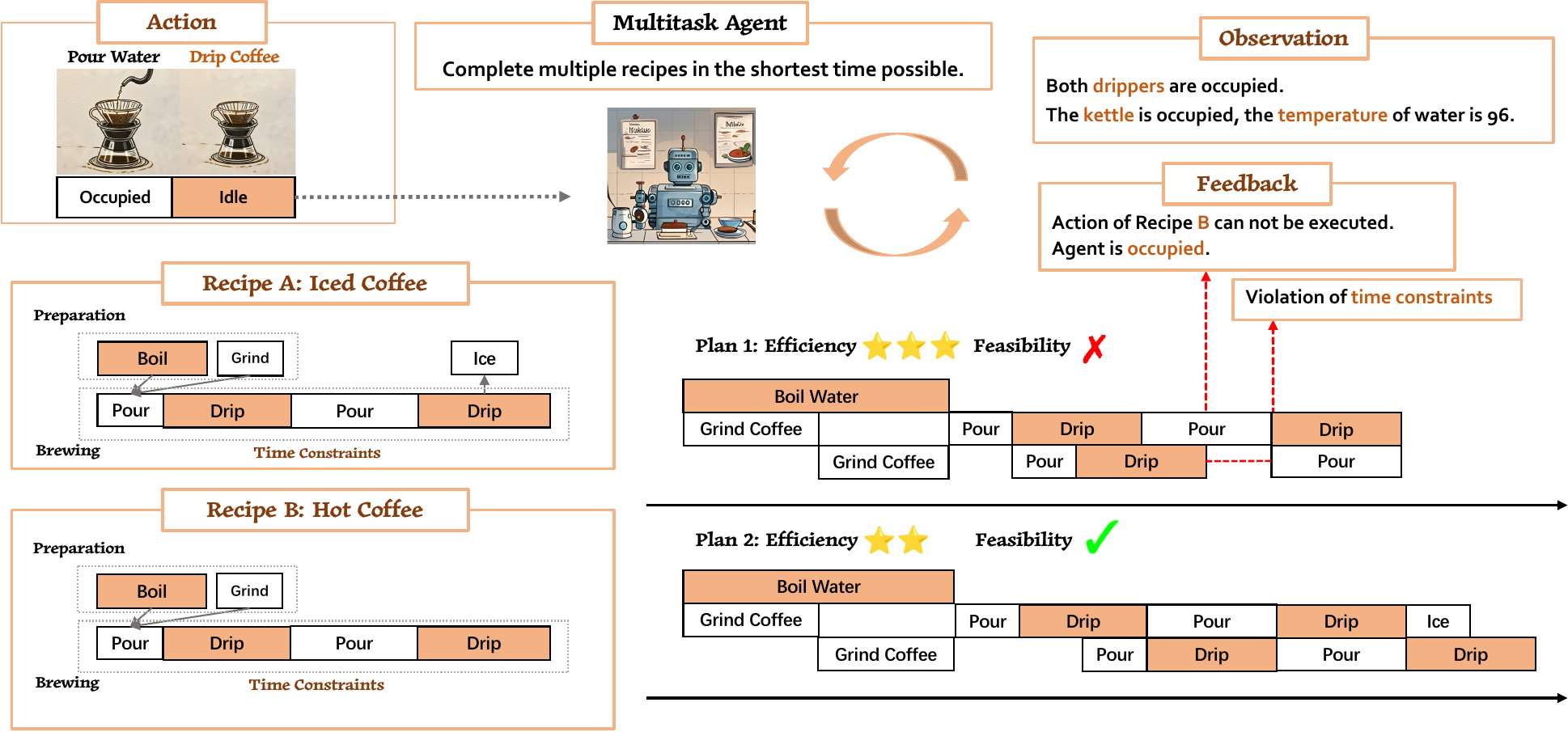}
    \caption{A simplified demonstration of our benchmark. Actions will either occupy the agent or leave it idle. 
    The four steps of brewing must be executed sequentially as time constraints between actions. The goal for the agent is to plan multitasking to complete the recipes in the shortest time possible without violating any constraints. The first plan illustrates a scenario where the agent attempts always to keep the agent occupied for higher efficiency, resulting in violations of time constraints. The second plan maintains the balance between the efficiency and feasibility of the plan by leaving the agent idle on purpose to maintain the time constraints for all actions. }
    \label{fig:intro}
\end{figure*}

The effectiveness of LLM-based agents is primarily evaluated based on the \textit{feasibility} of their plans in the scenarios of web browsing~\cite{yao2022webshop, zhou2023webarena}, tool usage~\cite{qin2023toolllm, li2023api}, computer manipulation~\cite{xie2024osworld, gou2024navigating} and agent navigation~\cite{shridhar2020alfworld,wang-etal-2022-scienceworld}.

However, the ability to manage concurrent objectives remains an often overlooked yet crucial requirement in real-world applications, as exemplified by everyday scenarios where humans prepare multiple dishes simultaneously for a meal or conduct parallel laboratory experiments
~\cite{russell2010artificial, zhang2024timearena, wu2024copilot}. Current planning benchmarks assume that models execute tasks by decomposing the overall goal into steps and achieving these subgoals sequentially, one at a time~\cite{liu2023agentbench,ma2024agentboard}. Consequently, these datasets fail to account for the duration of an action and the potential for multitasking. The multitasking scenario proposes a different objective in addition to \textit{feasibility}. It challenges the model to optimize the \textit{efficiency} to reach multiple goals simultaneously.


Time constraints between actions are often imposed in the recipe of dishes or experiments, indicating specific actions must be performed within a particular time interval after the preceding step is completed. For instance, the pouring and dripping actions for pour-over coffee must be carried out in sequence without any delay, as recommended by professionals~\cite{hoffmann2018world}. This property introduces a unique challenge for multitask planning apart from conventional benchmarks.
The first plan in Figure~\ref{fig:intro} illustrates that if the agent prioritizes maximizing efficiency by rushing to multitask whenever it is idle, it may inadvertently violate future time constraints. Consequently, the agent must balance the need for efficiency with adherence to time constraints to achieve feasible multitask planning as shown in the second plan in Figure~\ref{fig:intro}.


We propose a new benchmark \textsc{Recipe2Plan} based on real-world recipes and constraints to evaluate the multitasking abilities of agents. We highlight three main challenges as:
(1) \textbf{Commonsense Reasoning}: 
The agent must identify idle periods in the recipe as opportunities for multitasking, recognize action dependencies, and consider physical constraints to construct feasible action sequences.
(2) \textbf{Dynamic Local Planning}: 
As recipe states evolve based on executed actions, the agent must continuously determine executable actions at each timestep. 
Additionally, if the agent's initial assumptions about properties or constraints do not align with real-world conditions, it must dynamically adapt its beliefs and revise the plan accordingly. 
(3) \textbf{Strategic Global Planning}: The agent is required to allocate the use of physical objects and schedule actions on a timeline to enable efficient multitasking. It is crucial to avoid planning multitasking in a purely local and greedy manner, as this could lead to violations of time constraints. It challenges the agent to maximize efficiency while maintaining feasibility from a global perspective.

Our benchmark provides a testbed for the efficiency of LLM-based agents,
as they are approaching the upper limits of feasibility in current text-based agent benchmarks~\cite{sun2024adaplanner} and multitasking scenarios without time constraints between actions (Table~\ref{tab:main}). 
By introducing time constraints between actions, our benchmark evaluates the planning abilities of agents to maintain a delicate balance between efficiency and feasibility, rather than simply maximizing efficiency in a greedy manner. \textsc{Recipe2Plan} aims to push the boundaries of current agent planning capabilities, towards designing embodied agents that are capable of handling complex multitasking for industry-level tasks with time constraints between actions.
 
In this study, we experiment with various sizes of open-source models, such as Qwen2.5~\cite{qwen2.5} and Llama3.1~\cite{dubey2024llama}, as well as closed-source models, including Gemini-1.5-Pro~\cite{team2024gemini} and GPT-4o~\cite{openai2023gpt4}. 
Our experiments reveal that GPT-4o achieves the highest success rate of only 21.5\% and the main failure source is the violation of time constraints between actions. It suggests that current LLMs fail to deliver feasible plans while attempting efficient multitasking. We show that LLMs can deliver feasible plans if time constraints between actions are absent. However, their efficiency still lags significantly behind a simple heuristic method (\S\ref{sec3:experiments}). We also indicate that GPT-4o can trade efficiency for success rate if focusing solely on feasibility (\S\ref{subsec:priority}). Overall, we demonstrate that current LLMs struggle to balance efficiency and feasibility when multitasking with time constraints.
We further analyze the commonsense reasoning, local planning, and global planning capabilities of LLMs. By isolating each ability, we identify global planning as the primary source of task failure and inefficient multitasking (\S\ref{subsec:global_planning}). \looseness=-1

Our contributions are as follows:

\noindent$\bullet$ We introduce time constraints between actions to restrict the time interval between steps for multitask planning. This property highlights a new perspective to evaluate the planning abilities as balancing efficiency and feasibility compared with existing works focusing on feasibility solely.

\noindent$\bullet$ We construct a benchmark \textsc{Recipe2Plan} based on real-world recipes for multitask planning. It challenges the model to allocate the usage of physical objects and schedule actions on the timeline to complete the recipes in the shortest time possible without violating time constraints between actions.

\noindent$\bullet$ We evaluate open-source and closed-source models on our benchmarks. Our results show that LLMs struggle with planning multitasking under time constraints between actions, resulting in a low success rate for the task. This highlights the need for further development in temporal reasoning and global planning capabilities of LLM agents. 

\begin{table*}[t]
    \small
    \centering
    \begin{tabular}{lccccc}
    \toprule
        \textbf{Benchmark} & \textbf{\makecell{Commonsense \\Reasoning}} & \textbf{\makecell{Temporal \\ Planning}}& \textbf{\makecell{Multitask \\ Planning}} & \textbf{\makecell{Time  Constraints\\ Between Actions}} & \textbf{\makecell{Balance Efficiency \\ and Feasibility}} \\
        \midrule
        Graph Coloring~\cite{stechly2024self} & \textcolor{red}{\ding{55}} & \textcolor{red}{\ding{55}}&  \textcolor{red}{\ding{55}} & \textcolor{red}{\ding{55}} & \textcolor{red}{\ding{55}} \\
        \textsc{Natural Plan}~\cite{zheng2024natural} & \textcolor{red}{\ding{55}} & \textcolor{green}{\ding{51}}& \textcolor{red}{\ding{55}} & \textcolor{red}{\ding{55}} & \textcolor{red}{\ding{55}} \\
        TravelPlanner~\cite{xietravelplanner} & \textcolor{green}{\ding{51}}& \textcolor{green}{\ding{51}} & \textcolor{red}{\ding{55}} & \textcolor{red}{\ding{55}} &\textcolor{red}{\ding{55}} \\
        \textsc{TimeArena}~\cite{zhang2024timearena} &\textcolor{green}{\ding{51}}& \textcolor{green}{\ding{51}} & \textcolor{green}{\ding{51}} & \textcolor{red}{\ding{55}} & \textcolor{red}{\ding{55}} \\
        \midrule
        \textsc{Recipe2Plan} & \textcolor{green}{\ding{51}} & \textcolor{green}{\ding{51}}& \textcolor{green}{\ding{51}} & \textcolor{green}{\ding{51}} & \textcolor{green}{\ding{51}}\\
        \bottomrule
    \end{tabular}
    \caption{Comparison with existing \textit{scheduling} benchmarks. Two unique properties distinguish our benchmark: 
    (1) \textit{Time Constraints Between Actions}: Specific actions need to be performed within particular time intervals following the preceding steps. It reflects real-world scenarios where violating such constraints might lead to task failure or safety risk.
    (2) \textit{Balance Efficiency and Feasibility}: The introduction of time constraints presents a unique challenge where locally optimal planning for maximum efficiency might result in task failure. Therefore, the model must adopt a global perspective to balance efficiency and feasibility during planning.}
    \label{tab:comparison}
\end{table*}

\section{Related Work}
\noindent\paragraph{Planning Benchmarks.}
To evaluate the planning abilities of LLM-based agents, researchers have proposed benchmarks across various domains such as web browsing~\cite{yao2022webshop,zhou2023webarena,deng2024mind2web}, tool usage~\cite{qin2023toolllm,li2023api}, and computer manipulation~\cite{xie2024osworld,gou2024navigating}. These benchmarks assess an agent's ability to execute a sequence of actions to achieve a general goal in a partially observable environment~\cite{liu2023agentbench,ma2024agentboard}. However, these environments do not account for the duration of each action. Additionally, they evaluate planning abilities based solely on feasibility, without comparing the efficiency of task completion between different agents.

\noindent\paragraph{Scheduling Benchmarks.}
Apart from the typical planning task in which the agent interacts with a partially observable environment without prior knowledge of how to achieve the goal, the scheduling task provides the agent with a complete description of the task. The objective is to deliver an action sequence from a small set of fixed actions to meet the given objectives~\cite{pinedo1992scheduling, smith2000bridging, valmeekam2024planning}.
Graph coloring~\cite{stechly2024self} investigates whether LLMs can self-critique their answers for violations of scheduling constraints. \textsc{NaturalPlan} assesses scheduling abilities in contexts such as trip planning, meeting planning, and calendar scheduling. TravelPlan~\cite{xietravelplanner} deals with more complex commonsense constraints and strict restrictions. 
\textsc{TimeArena} primarily evaluates the multitasking capabilities of LLMs in scenarios without time constraints between actions. In contrast, our work focuses on assessing the ability to balance efficiency and feasibility under time constraints between actions. While \textsc{TimeArena} only analyzes task failures arising from a mix of commonsense reasoning, local planning, and global planning, our work conducts extensive analysis to isolate each challenge, identifying global planning as the primary bottleneck. This systematic analysis provides deeper insights into the complexities of time-sensitive multitasking.

\noindent \paragraph{Planning Methods.}
Different methods use feedback and instructions in various ways. Open-loop methods such as Chain-of-Thought~\cite{wei2022chain}, least-to-most~\cite{zhou2022least} and plan-and-solve~\cite{wang-etal-2023-plan} plan the action sequence without any feedback from the environment. This type of method is vulnerable to the hallucination of execution constraints and environment dynamics. Closed-loop methods such as  ReAct~\cite{yao2022react} and Reflexion~\cite{shinn2024reflexion} only refine local actions, which might result in global failure due to time constraints. AdaPlanner~\cite{sun2024adaplanner} refines the entire plan based on environmental feedback and past failures. 
\section{\textsc{Recipe2Plan}}

\textsc{Recipe2Plan} evaluates the planning ability of LLMs for efficient and feasible multitasking under constraints. Specifically, we provide the model with multiple goals that can be achieved by following recipes. Each recipe $\boldsymbol{A}$ is represented as a linear sequence of actions $\boldsymbol{A}=(a_{0}, a_{1},..., a_{n})$, with each action assigned a specific execution time $t_{n}$. The task is to plan the action sequence to complete all goals in the shortest time possible, adhering to the properties and constraints detailed in \S\ref{sec:property} and \S\ref{sec:constraints}. The statistics of our benchmark are presented in Table~\ref{tab:statistics}. \textsc{Recipe2Plan} challenges the model to apply commonsense reasoning to infer any unwritten constraints from the recipe, including action concurrency, action dependencies, and resource limitations while planning the action sequence to minimize overall execution time.

\subsection{Properties of Actions} 
\label{sec:property}
\noindent\paragraph{Action Duration.} This refers to the time required for an agent to complete a specific action. For the coffee recipes illustrated in Figure~\ref{fig:intro}, the duration of actions such as pouring and dripping is fixed, and any deviation from these durations can result in spoiled flavor. Following this principle, each action in our benchmark is annotated with a specific duration. The recipe in the dataset explicitly states this duration, allowing the agent to accurately schedule the timeline. We also allow agent to pause interruptable actions for more flexible multitasking.

\noindent\paragraph{Action Concurrency.} Continuous actions, such as \textit{pour water}, require the active involvement of the agent while the action is in progress. In contrast, autonomous actions, like \textit{boil water}, do not require the agent's continuous attention, allowing the agent to remain idle and free to perform other tasks concurrently. Identifying autonomous actions and executing them simultaneously with other actions is the key to efficient multitasking.


\subsection{Multitasking Constraints}
\label{sec:constraints}
\noindent\paragraph{Action Dependencies.} The dependent relationships between actions are generally not explicitly stated in the recipe. Although the actions in a recipe are often presented in a linear sequence, the action dependencies might form a graph structure. For example, as illustrated in Figure~\ref{fig:intro}, step 3 \textit{pour water} depends on step 1 \textit{boil water} and step 2 \textit{grind coffee}, but steps 1 and 2 can be performed independently of each other. 
This property challenges the agent to dynamically identify the executable actions at each timestep as the recipe status evolves.

\noindent\paragraph{Resource Limitations.} During planning, the agent must recognize whether an object is occupied at the current time and when it will be available again. Different recipes may require different physical objects and conditions. As shown in Figure~\ref{fig:intro}, the pouring process for coffee requires water at a specific temperature, so the agent can boil water for both recipes simultaneously to speed up the process. However, if different recipes require water or an oven at different temperatures, the agent must sequentially prepare the object for each recipe based on when it becomes available. This property necessitates that the agent plan globally, scheduling the use of different objects while considering the duration of actions and specific condition requirements.\looseness=-1

\noindent\paragraph{Time Constraints Between Actions.} This property is crucial for feasible multitasking of professional coffee preparation~\cite{hoffmann2018world} and delicate biomedical experiments~\cite{itoh2021optimal}, where specific actions must be executed within a precise time interval following a preceding action. Failure to adhere to these time constraints may cause the entire recipe or experiment to fail. This realistic property imposes a significant challenge on multitask planning. As depicted in Figure~\ref{fig:intro}, the agent can not simply follow a greedy manner that prioritizes immediate actions without considering the broader temporal constraints. 
Incorporating time constraints between actions ensures that the agent must strategically balance multitasking efficiency with feasibility.

\subsection{Dataset Construction}

\noindent\paragraph{Recipe Annotation.} 
We collect and clean recipes with annotated dependent relations from a website for cooking\footnote{\url{www.instructables.com}}. We ask three annotators to label the properties and constraints following the pipeline in Appendix~\ref{subsec:recipe_collection}.
The average kappa scores among annotators are 0.78 for action concurrency, 0.50 for action interruptibility, 0.66 for time constraints,  and 0.86 for resource limitations. Based on these results, we explicitly list action interruptibility and time constraints in the recipes, while keeping action concurrency and resource limitations as implicit properties that agents need to identify through commonsense reasoning.
Finally, we prompt GPT-4o~\cite{hurst2024gpt} to annotate the durations and time intervals that are also provided in the recipes. All annotators have reviewed these annotations to ensure their reasonableness. \looseness=-1

\noindent\paragraph{Combine Recipes for Multitasking.}
We carefully select recipe combinations to evaluate planning abilities for efficient multitasking. To keep the action space and context length manageable, we only combine two recipes at a time. We then adapt a heuristic algorithm from \citet{zhang2024timearena} to plan action sequences for multitasking. Multitasking efficiency for each sequence is computed according to Equation~\ref{equation:me}. Instances are chosen for the benchmark based on the following criteria:
(1) Opportunities for multitasking: We include instances with multitasking efficiency higher than 80\% when planning without time constraints, indicating significant potential opportunities for multitasking.
(2) Balance of efficiency and feasibility: We select instances in which the multitasking efficiency drops when time constraints between actions are considered, suggesting that an efficient greedy planning strategy would likely violate these time constraints.
(3) Sufficient benchmark size with reasonable cost: We include 65 diverse recipe combinations to reach a similar size to previous work TimeArena.

\subsection{Environment}
We implement an environment to provide feedback to the agent. The agent can choose to perform one action for a specified duration at a given time. If the agent determines that no action can be performed at the moment, it can choose the time for its next planned action. The environment then receives the action and checks for any constraint violations. If a constraint is violated, the environment will specify the type of violation. If the action is permissible, feedback from the environment includes the status of physical objects, completed actions, and ongoing autonomous actions. We present examples of observations and feedback in Appendix~\ref{appsub:environment}. The agent can use this feedback to revise its global plan and decide on the next action.

\begin{table}[t]
    \centering
    \small
    \begin{tabular}{lc}
    \toprule
    \multicolumn{2}{c}{\textbf{Recipe Statistics}} \\
    \midrule
    \# Recipes & 29 \\
    Avg. Actions per Recipe &13.1 \\
    Avg. Autonomous Actions per Recipe&  3.4  \\
    Avg. Interruptible Actions per Recipe&  3.9 \\ 
    Avg. Time constraint Between Actions & 3.1 \\
    Avg. Duration per Action (min) &  5.7  \\
    Avg. Restriction Interval (min) & 2.7   \\
    \midrule
    \multicolumn{2}{c}{\textbf{Multitasking Statistics}} \\
    \midrule
    \# Instances &  65\\
    Avg. Executable Action per Step & 3.1 \\
    Avg. Efficiency w/o time constraint (\%) & 86.1 \\
    Avg. Efficiency w/ time constraint (\%) & 73.7 \\
    \bottomrule
    \end{tabular}
    \caption{Statistics of recipes and multitasking instances in our \textsc{Recipe2Plan} benchmark.  The agent can choose any timestamp for the next action, expanding the search space beyond the number of executable actions solely. }
    \label{tab:statistics}
\end{table}
\section{Experiments}
\label{sec3:experiments}
\subsection{Baselines}
\noindent \textbf{Models.} We evaluate several models, including the open-source Llama-3.1 with parameter sizes of 8B and 70B~\cite{dubey2024llama}, and Qwen2.5 with parameter sizes of 7B, 32B and 72B~\cite{qwen2.5}. Additionally, we assess the closed-source models, GPT-4o-mini, GPT-4o~\cite{hurst2024gpt} and Gemini-1.5-Pro~\cite{team2024gemini}. The versions of the models are detailed in Appendix~\ref{appsub:model_version}.

\noindent{\textbf{Methods.}} We begin by prompting the model to identify any unwritten properties and constraints from each recipe. These identified elements are then concatenated with the original description. Next, we employ a \textit{ReAct}-style prompting method~\cite{yao2022react} on the models to plan the action sequence. To evaluate the planning abilities and mitigate the cascading errors from commonsense reasoning, we also experiment with an oracle setting \textit{ReAct + Oracle} that replaces the identified constraints with the gold annotations.

\noindent \textbf{Constraint Setting.} We evaluate the agent under \textit{without time constraints} and \textit{with time constraints} settings to study the impact of time constraints between actions on the feasibility and efficiency of multitasking agents.

\begin{table*}[t]
\centering
\small
\begin{tabular}{lcccccccccc}
\toprule
 \multirow{2}{*}{\textbf{Model}} & \multicolumn{4}{c}{\textbf{w/o Time Constraints}}  & \multicolumn{4}{c}{\textbf{w/ Time Constraints}} \\
\cmidrule(lr){2-5} \cmidrule(lr){6-9} 
& \multicolumn{1}{c}{\textbf{Success} }  & \multicolumn{1}{c}{\cellcolor{white}\textbf{Progress} }&   \multicolumn{1}{c}{\textbf{R-Efficiency} } & \multicolumn{1}{c}{\textbf{S$\times$E} } & \multicolumn{1}{c}{\textbf{Success}  }&  \multicolumn{1}{c}{\textbf{Progress}  }&  \multicolumn{1}{c}{\textbf{R-Efficiency}  } & \multicolumn{1}{c}{\textbf{S$\times$E} } \\
\midrule
& \multicolumn{8}{c}{\textit{ReAct}}\\
\midrule
\textit{Open-Source Models} & & & & & & & & \\
\quad Qwen2.5-7B &1.5 & 26.5 & 73.6 & 1.3 & 0.0 & 22.4 & 78.0 & 0.0 \\
\quad Llama-3.1-8B &0.0 & 9.7 & 65.2 & 0.0 & 0.0 & 10.3 & 59.7 & 0.0 \\
\quad Qwen2.5-32B & \underline{80.0} & \underline{96.7} & 57.3 & 47.2 & \underline{15.4} & \underline{57.4} & 91.4 & \underline{8.6} \\
\quad Llama-3.1-70B &72.3 & 88.8 & 66.5 & 48.7 & 13.8 & 55.5 & 95.9 & 10.4 \\
\quad Qwen2.5-72B &72.3 & 91.2 & 71.0 & \underline{48.8} & 7.7 & 54.9 & 97.1 & 4.5 \\
\textit{Closed-Source Models} & & & & & & & & \\
\quad GPT-4o-mini &3.1 & 51.8 & 63.5 & 1.8 & 1.5 & 36.0 & 68.3 & 0.4 \\
\quad Gemini-1.5-Pro & 20.0 & 66.4 & \underline{76.7} & 14.8 & 3.1 & 47.5 & \underline{100.2} & 1.8 \\
\quad GPT-4o &\textbf{90.8} &\textbf{ 99.1} & \textbf{78.2} & \textbf{72.2} & \textbf{21.5} & \textbf{64.0} & \textbf{109.5} & \textbf{20.5} \\
\midrule
& \multicolumn{8}{c}{\textit{ReAct + Oracle}}\\
\midrule
\textit{Open-Source Models} & & & & & & & & \\
\quad Qwen2.5-7B &0.0 & 28.1 & 90.7 & 0.0 & 0.0 & 24.1 & 73.9 & 0.0 \\
\quad Llama-3.1-8B &0.0 & 10.8 & 76.3 & 0.0 & 0.0 & 10.0 & 59.7 & 0.0 \\
\quad Qwen2.5-32B & \underline{80.0} & \underline{96.7} & 61.2 & \underline{51.0} & \underline{10.8} & \underline{57.0} & 91.7 & \underline{8.6} \\
\quad Llama-3.1-70B &73.8 & 89.0 & 67.5 & 49.6 & 6.2 & 52.9 & 89.6 & 2.5 \\
\quad Qwen2.5-72B &72.3 & 90.1 & 70.2 & 50.7 & 7.7 & 52.6 & \underline{106.5} & 7.0 \\
\textit{Closed-Source Models} & & & & & & & & \\
\quad GPT-4o-mini &10.8 & 55.8 & 61.5 & 3.8 & 1.5 & 35.0 & 82.3 & 1.8 \\
\quad Gemini-1.5-Pro & 16.9 & 63.6 & \textbf{82.8} & 14.1 & 7.7 & 49.1 & 105.1 & 4.4 \\
\quad GPT-4o &\textbf{95.4} &\textbf{ 99.4} & \underline{75.0} & \textbf{73.1} & \textbf{27.7} & \textbf{60.6} & \textbf{116.0} & \textbf{24.0} \\
\midrule
Heuristics  & 100.0 & 100.0 & 100.0 & 100.0 & 100.0 & 100.0 & 100.0 & 100.0 \\

\bottomrule
\end{tabular}
\caption{Results of \textit{ReAct} and \textit{ReAct + Oracle} experiments on \textsc{Recipe2Plan}. We report average percentage of success rate (\textbf{Success}), progress rate (\textbf{Progress}), relative multitask efficiency (\textbf{R-Efficiency}) and muititasking score (\textbf{S$\times$E}). \textbf{Bold} denotes the best performance and \underline{underline} denotes the second-best performance.}
\label{tab:main}
\end{table*}

\subsection{Evaluation Metrics}

\noindent \textbf{Success Rate.} It measures the \textit{feasibility} of the plans exclusively by evaluating whether the agent can deliver a plan that successfully completes all recipes.  The agent might fail due to a violation of time constraints, reaching a maximum of execution errors, or being stuck in an endless dead loop.


\noindent \textbf{Progress Rate.} This metric measures the proportion of successfully executed actions in the recipes. It evaluates the \textit{feasibility} of the planning process with a more fine-grained perspective.



\noindent\textbf{Multitasking Efficiency.} Autonomous actions within recipes enable time savings, quantified as $T_{save}$
 , relative to the total duration of completed actions. We define multitasking \textit{efficiency} as the ratio of saved time to the cumulative duration of autonomous actions $T_{auto}$, with detailed rationale in Appendix~\ref{appsec:efficiency}.

\begin{equation}
    \operatorname{Efficiency}_{agent}  = \frac{T_{save}}{T_{auto}}
    \label{equation:me}
\end{equation}

\begin{figure}[h]

\centering

\includegraphics[width=0.8\linewidth]{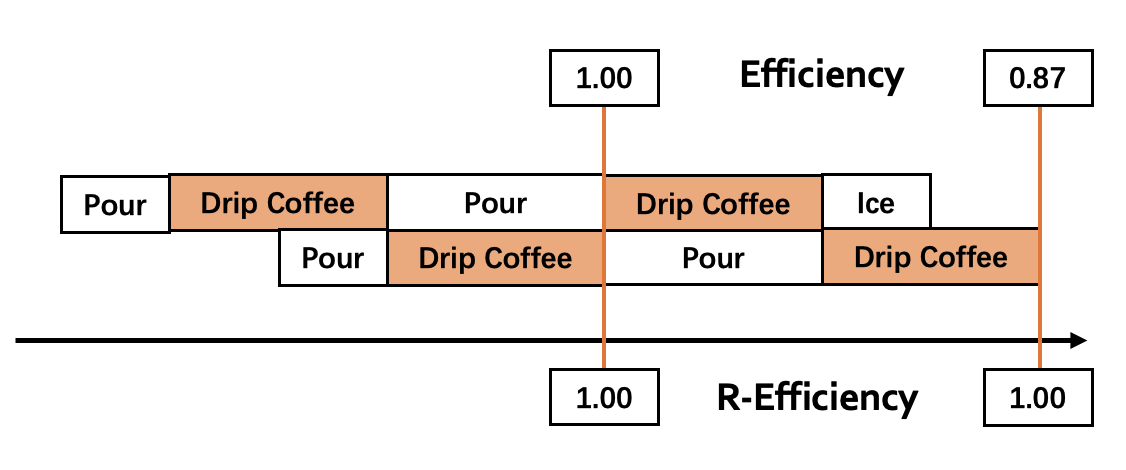}

\caption{Demonstration of relative efficiency. The efficiency is affected by the progress rate and we use relative efficiency (R-Efficiency) to calibrate the metric.}

\label{fig:relative_efficiency}

\end{figure}

The efficiency metric is influenced by the progress rate. As shown in Figure~\ref{fig:relative_efficiency}, if the agent aborts the interaction midway, it will achieve an efficiency of 100\% while feasible and optimal manner only obtains an efficiency of 87.5\%. To address this bias, we adjust the heuristic search plan to match the agent's progress rate and compute its efficiency as 
$\operatorname{Efficiency}_{ref}$. For example, if an agent completes 5 steps, we calculate the efficiency of the first 5 steps of the heuristic plan as the reference.

The relative efficiency is computed as:

\begin{equation}
        \operatorname{R-Efficiency}  = \frac{\operatorname{Efficiency}_{agent}}{\operatorname{Efficiency}_{ref}}
    \label{equation:rme}
\end{equation}

\noindent\textbf{Multitasking Score.}  We use this metric to present the overall \textit{efficiency} and \textit{feasibility} of the plans simultaneously. The score is computed as:
\begin{equation}
        \operatorname{M. Score} = \left\{\begin{array}{ll}
\operatorname{R-Efficiency} & \operatorname{Success}  = 1 \\
0 & \text {otherwise}
\end{array}\right.
\end{equation}
The rationale is that the agent should prioritize ensuring the successful completion of the recipes before aiming to achieve higher efficiency in task execution. The overall score is computed as the average of multitasking scores for each instance.

\begin{figure}[t]

    \centering
    \includegraphics[width=0.8\linewidth]{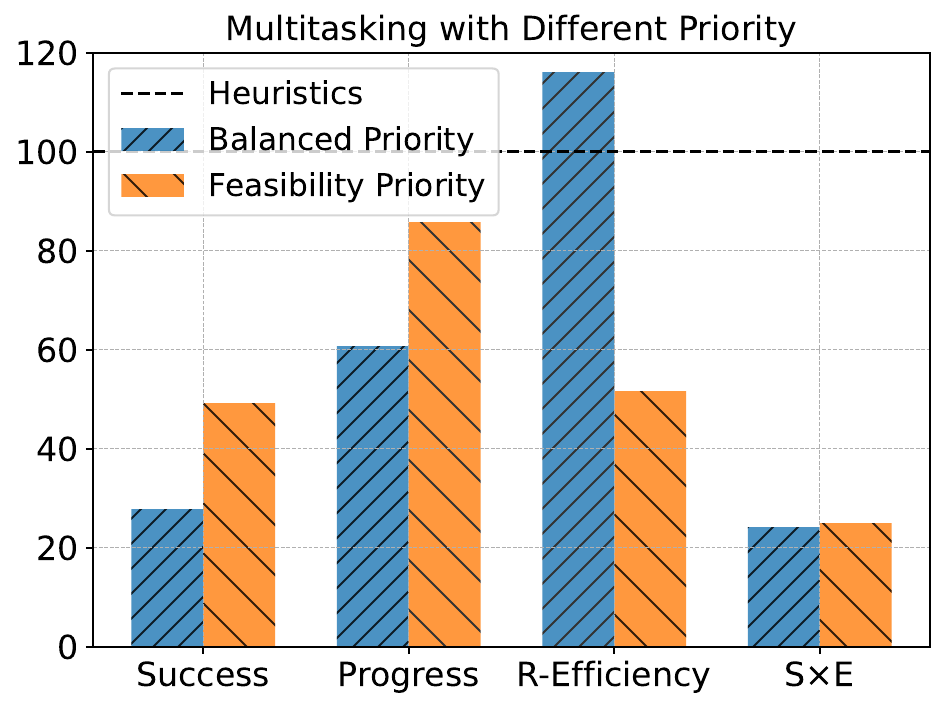}
    \caption{Results of GPT-4o planning with different priority. \textit{Balanced Priority}: Blend feasibility and efficiency as in \S\ref{sec3:experiments}. \textit{Feasibility Priority}: Only focus on feasibility without considering efficiency. }
    \label{fig:priority}
\end{figure}

\subsection{Main Results}
\noindent \textbf{LLMs can plan feasible multitasking in the absence of time constraints between actions, but efficiency needs improvement.} 
In the \textit{ReAct} setting, GPT-4o delivers 90.8\% feasible plans for multitasking, surpassing other tested models by a large margin. This demonstrates its ability to revise beliefs about unwritten properties and constraints and to correct its actions to complete tasks. GPT-4o achieves a multitasking efficiency of 78.2\%, indicating that there is still room for improvement in multitasking efficiency with LLMs. 

\noindent \textbf{LLMs face challenges in balancing efficiency and feasibility for multitasking under time constraints between actions.} The success rates and completion ratios of all models decrease significantly when multitasking with time constraints. GPT-4o only achieves the highest success rate of 21.5\%. Interestingly, Gemini-1.5-Pro and GPT-4o achieve a relative efficiency higher than the heuristic baseline. This indicates a tendency to prioritize high efficiency during local planning. However, the agent fails to maintain feasibility for time constraints from a global perspective while managing multitasking efficiency.


\noindent \textbf{Commonsense reasoning is not the bottleneck for feasible and efficient multitasking. The challenge lies in effectively leveraging constraints to schedule the actions.}
The models generally achieve F1 scores higher than 70\% for commonsense reasoning as detailed in Appendix~\ref{appsec:commonsense}. 
We investigate the impact of unidentified properties and constraints in \textit{ReAct + Oracle} setting. While the success rate improves by 4.8\% for GPT-4o in the \textit{w/o time constraints} setting, the relative efficiency for GPT-4o decreases by 2.8\%, indicating that the model struggles to formulate an efficient multitasking plan even with oracle constraints. In the \textit{w/ time constraints} setting, we observe the success rate increases slightly for GPT-4o from 21.5\% to 27.7\%. Oracle constraints also improve efficiency for Qwen2.5-72B and Gemini-1.5-Pro, surpassing the heuristic baseline, yet their success rates remain below 10\%. This suggests that misidentified constraints are not the primary cause of time constraint violations. Instead, the results highlight that models struggle to fully leverage explicit constraints to plan efficient and feasible multitasking.


\begin{figure}[t]

    \centering
    \includegraphics[width=0.98\linewidth]{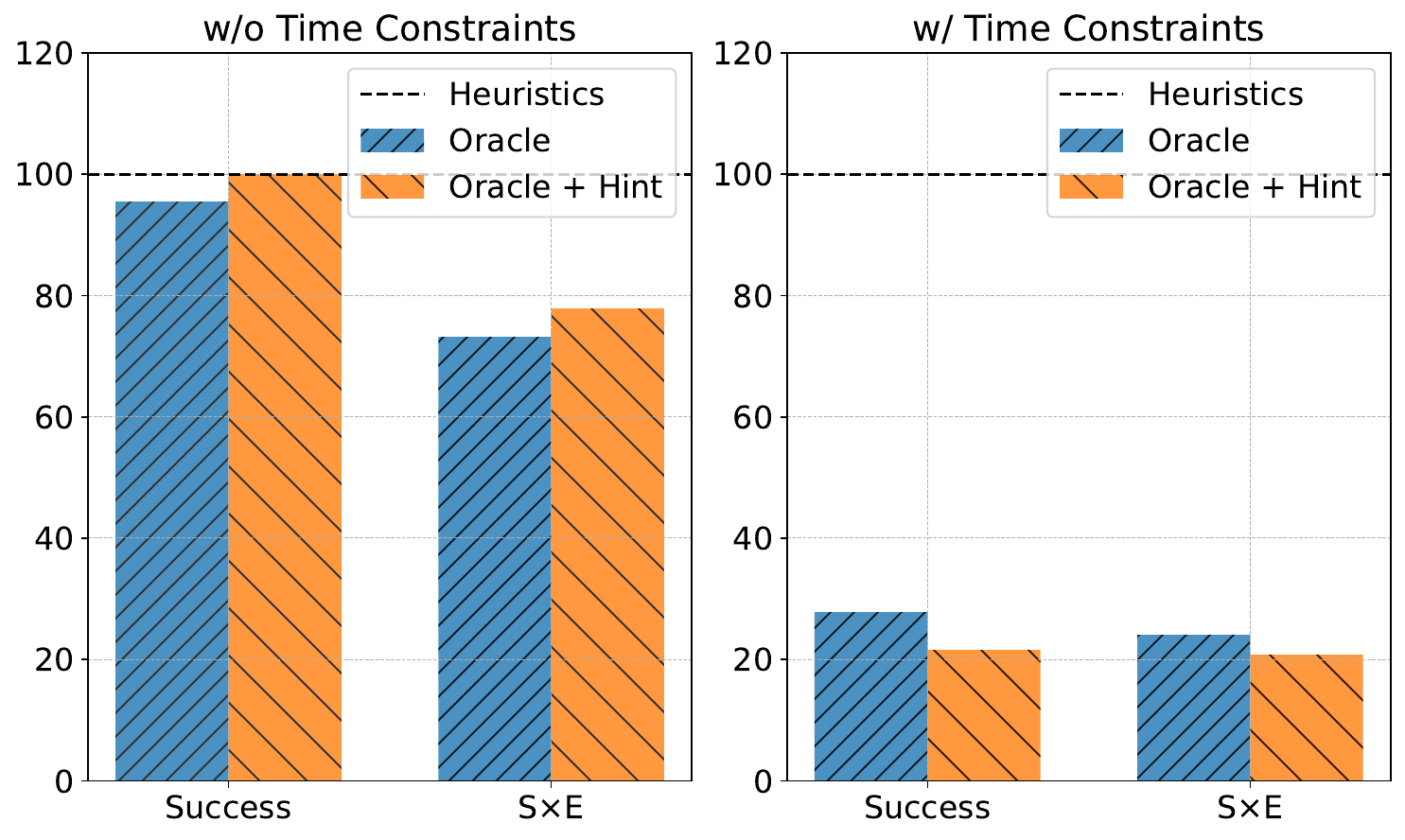}
    \caption{Results of prompting GPT-4o under \textit{Oracle} setting: gold constraints, and \textit{Oracle + Hint} setting: gold constraints and executable actions at each step. }
    \label{fig:hint}
\end{figure}

\begin{figure*}[t]
    \centering
    \begin{subfigure}
        \centering
        \includegraphics[width=0.42\linewidth]{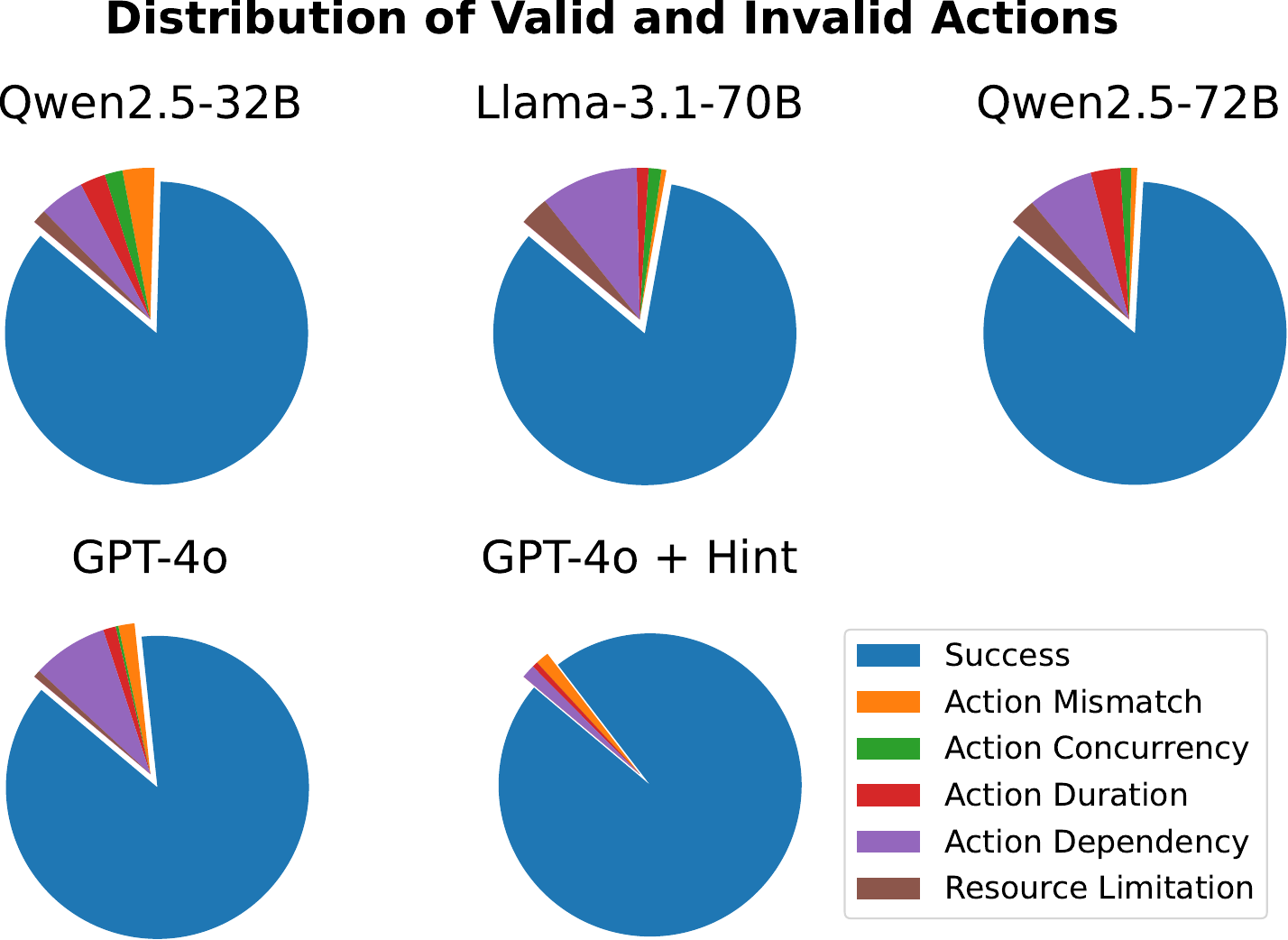}
    \end{subfigure}
    \quad
    \begin{subfigure}
        \centering
        \includegraphics[width=0.42\linewidth]{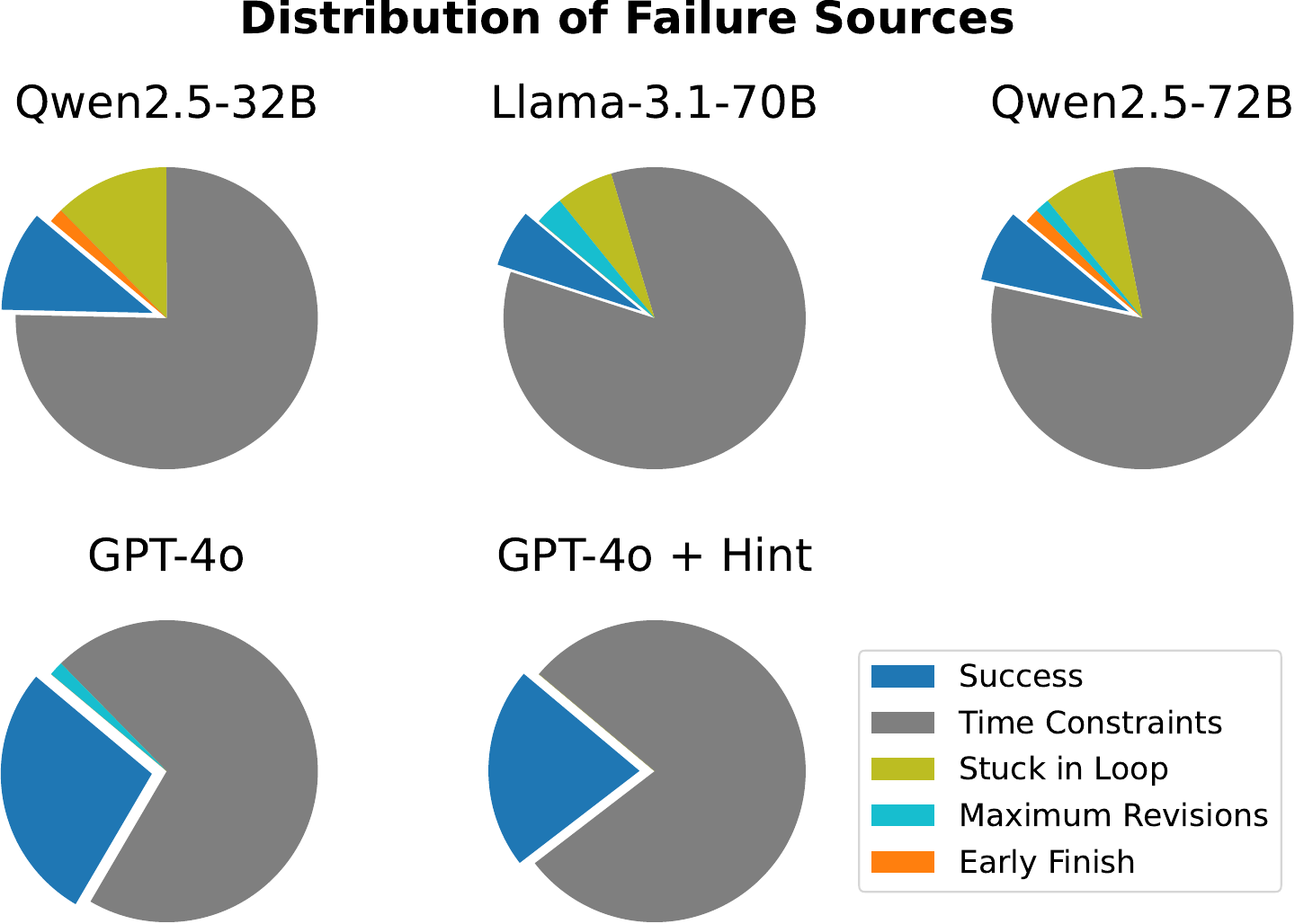}
    \end{subfigure}
    \caption{Analysis of the distribution of invalid actions and failure source of \textit{ReAct + Oracle} agents planning with time constraints. GPT-4o + Hint: We add all the executable actions in the prompt to help the agent choose the next action during dynamic local planning.}
    \label{fig:error}
\end{figure*}

\section{Analysis}
\subsection{Multitasking with Different Priority}
\label{subsec:priority}
Our experiments (\S\ref{sec3:experiments}) indicate that LLMs struggle to complete the recipes with \textit{balanced priority} of feasibility and efficiency. Therefore, we evaluate if LLMs can focus solely on the \textit{feasibility priority}, ensuring that recipes are completed without violating any constraints as detailed in Appendix~\ref{appsub:method_details}.\looseness=-1

\noindent \textbf{LLMs can ensure more task completion by trading efficiency for feasibility.} The results in Figure~\ref{fig:priority} show that under the feasibility priority setting, the success rate significantly increases from 27.7\% to 49.2\%, and the progress rate increases from 60.6\% to 85.7\%. This indicates that focusing on feasibility allows more recipes to be completed and more steps to be executed within the given time constraints. It further underscores the importance of enhancing the planning abilities of LLM agents to balance feasibility and efficiency.


\subsection{Error Analysis}
In this section, we take a closer look at the dynamic local planning abilities of the agents by examining the distribution of valid and invalid actions. Invalid actions are categorized into: \textit{action mismatch} (executing non-existing actions and repeating finished actions), and violations of other properties and constraints. We distinguish \textit{time constraint} as a source of failure separate from invalid actions, along with other types of failure in Figure~\ref{fig:error}.

\noindent \textbf{Action dependencies are the primary source of invalid actions.} As illustrated in Figure~\ref{fig:error}, models with a high success rate under the \textit{w/o time constraints} setting consistently achieve a valid action ratio above 80\% under time constraints. Despite all constraints being explicitly presented in the prompt during planning, the agent frequently violates these constraints, particularly those related to action dependencies. Upon examining the reasoning traces, we observe that LLMs often breach action dependencies constraints while attempting to optimize multitasking, consequently neglecting feasibility.

\noindent \textbf{Time constraints between actions are the main sources of task failure.} For the failure source of planning with time constraints, open-source models Llama-3.1 and Qwen2.5 may still get stuck in loop or exceed maximum revisions for about 10\% of the instances. But the main source for the failure of planning is due to time constraints between actions, even GPT-4o fails to maintain time constraints in 70\% of the cases.

\subsection{Planning with Hints of Executable Actions}
\label{subsec:global_planning}
 As LLMs can not handle action dependencies well while planning for efficient multitasking, we further add the executable actions for each step in the prompt.  This allows us to evaluate global planning abilities directly as hints significantly improve the model’s local planning ability to choose valid next steps (Figure~\ref{fig:error}). 

\noindent \textbf{LLMs lack global planning ability for efficient planning and maintaining time constraints between actions.}
The success rate and multitask score show minimal improvement in both settings, as illustrated in Figure~\ref{fig:hint}. It suggests that agents fail to consider the impact of executable actions on the overall feasibility and efficiency from a global perspective. 
Table~\ref{tab:case_wo_tc} in the Appendix demonstrates a case where GPT-4o fails to estimate the priority of autonomous actions and leaves the agent idle during the execution of the last two actions.
Table~\ref{tab:case_w_tc} in the Appendix provides an example where GPT-4o rushes to \textit{heat up oil} at the beginning of the plan and executes this action concurrently with others to maximize efficiency. This plan overlooks ingredient preparation and results in the oil heated for an extended period. It does not only violate the time constraint but also risks catching fire.


\section{Conclusions}
Our paper introduces the \textsc{Recipe2Plan} benchmark, which evaluates the feasible and efficient multitasking abilities of existing LLMs. This benchmark pushes the limits of current agent planning capabilities beyond mere task completion to include the optimization of time and resource management. Our experiments reveal that while strong models like GPT-4o can generate feasible plans without time constraints, their performance decreases sharply when time constraints are imposed. This highlights a significant gap between current capabilities and the requirements for feasible and efficient multitasking. Our analysis identifies global planning as the primary area needing improvement, paving the way for future work to focus on enhancing temporal reasoning and strategic planning. 

\section*{Limitations}
While multitasking is a practical application for LLM agents, our text-based environment does not fully capture the complexities of real-world cooking and experimentation. Our agent does not engage in physical exploration or interact with objects in the real world, focusing solely on the temporal planning aspects of multitasking. In our setting, the agent is assumed to perform every action without delay or failure. Introducing scenarios where the agent must search for ingredients in a kitchen or lab similar to \citet{shridhar2020alfworld} and \citet{wang-etal-2022-scienceworld} would present a more realistic and challenging environment. We plan to implement such a realistic environment in future work.


The metric we use to evaluate efficiency by computing the speed of completion may be biased by the progress rate. To address this, we introduce a relative multitasking efficiency metric to calibrate our evaluation. However, the solution provided by our heuristic baseline does not guarantee the optimal plan for the task. The search space is complex because the model can choose to execute actions at arbitrary time stamps and split actions into arbitrary time intervals, making it beyond the scope of classical scheduling algorithms with time constraints~\cite{itoh2021optimal}. While existing scheduling algorithms may take a long time to execute, our heuristic algorithm quickly identifies a feasible and efficient plan, though it may be suboptimal. We believe this heuristic can still serve as a valuable baseline for evaluating the multitasking abilities of agents. For future work, we plan to explore scheduling algorithms that can better handle the complexities of multitasking with time constraints.

\section*{Acknowledgement}
We thank Chang Ma for helpful suggestions on the experiment design.

\bibliography{custom}

\appendix

\label{sec:appendix}

\begin{table*}[t]
    \small
    \centering
    \begin{tabularx}{\textwidth}{m{1.0\columnwidth}m{1.0\columnwidth}}
    \toprule
{\#\# Recipe 1:Tacos\newline
Step 0 (3 min): Place the fish in a cooking pot with enough water to cover them.\newline
Step 1 (20 min): Let it boil for around 20 minutes.\newline
Step 2 (3 min): Drain them and put them in a bowl.\newline
Step 3 (3 min): Use a fork to smash them.\newline
Step 4 (5 min): Chop the onion and tomato in little squares.\newline
Step 5 (2 min): Put the onion in a pan with a little oil.\newline
Step 6 (5 min): Let it cook a little for 5 minutes.\newline
Step 7 (2 min): Place the tomato and stir.\newline
Step 8 (5 min): Let it cook for 5 minutes stirring constantly.\newline
Step 9 (5 min): When it's ready put the fish in the pan and mix well.\newline
Step 10 (1 min): Add salt and pepper to taste.\newline
Step 11 (15 min): Let it cool down.\newline
Step 12 (1 min): Warm the tortillas for 1 minute in the microwave.\newline
Step 13 (2 min): Put the fish we made in the middle of the tortilla using a spoon and make sure it doesn't reach the edge.\newline
Step 14 (2 min): Fold the tortilla in half and put one toothpick on each side to hold it closed.\newline
Step 15 (10 min): Fry them in a pan with oil\newline
Step 16 (2 min): Take the toothpicks off and serve when they are still warm.\newline\newline
Interrutable steps: 3, 4.\newline
Autonomous actions: step 1, 6, 11, 12, 15.\newline
Action Dependency: 0->1, 1->2, 2->3, 4->5, 5->6, 6->7, 7->8, 3->9, 8->9, 9->10, 10->11, 12->13, 11->13, 13->14, 14->15, 15->16.\newline
Steps 1, 6, 7, 8, 9, 10, 15 require stove, Steps 12 requires microwave.} &{\#\# Recipe 2:Smore-Bars  \newline                   
Step 0 (10 min): Preheat your oven to 350 degrees fahrenheit.\newline
Step 1 (3 min): Grease a 9x13 inch pan. \newline
Step 2 (1 min): Melt your 1 cup of butter in the microwave until it is completely melted.\newline
Step 3 (5 min): crush 2 cups (approximately 2 sleeves) of graham crackers. \newline
Step 4 (3 min): Mix the melted butter and crushed graham crackers together. \newline
Step 5 (5 min): Take about 3/4 (doesn't need to be exact) of your butter/graham cracker mixture and press into the bottom of your greased pan.\newline
Step 6 (2 min): Unwrap your candy bars and arrange them.  \newline
Step 7 (3 min): Evenly spread out your bag of mini marshmallows across entire pan. \newline
Step 8 (2 min): Sprinkle your remaining butter/graham cracker mixture across pan. \newline
Step 9 (15 min): Place pan in oven for 15 minutes.\newline
Step 10 (2 min): Cut and Enjoy!\newline
Interrutable steps: 1, 3, 5, 6, 7, 8, 10.\newline
Autonomous actions: step 0, 2, 9.\newline
Action Dependency: 2->4, 3->4, 1->5, 4->5, 5->7, 6->7, 7->8, 0->9, 8->9, 9->10.\newline
Steps 0, 9 require oven, Steps 2 requires microwave.
}\\
\multicolumn{2}{c}{
    \includegraphics[width=0.9\textwidth]{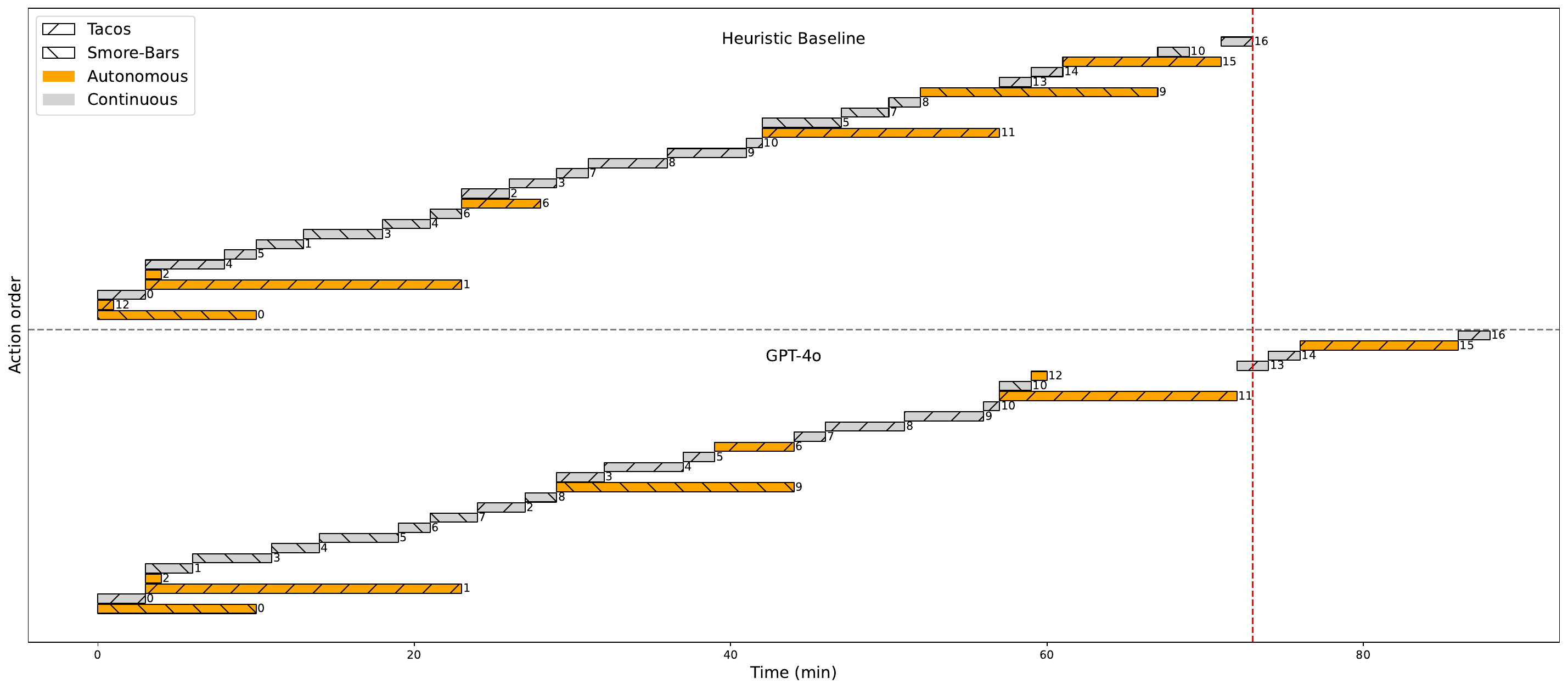}
}\\
\bottomrule 
\end{tabularx}
    \caption{Case study of GPT-4o planning without time constraints. GPT-4o result in a lower efficiency compared with the heuristic baseline. The primary difference is that GPT-4o prioritizes different autonomous actions and leaves the agent idle during the execution of the last two actions}
    \label{tab:case_wo_tc} 
\end{table*}

\begin{table*}[t]
    \small
    \centering
    \begin{tabularx}{\textwidth}{m{1.0\columnwidth}m{1.0\columnwidth}}
    \toprule
{\#\# Recipe 1: Vada\newline
Step 0 (5 min): First, start with blending the chilies, ginger and coriander along with some cumin seeds.\newline
Step 1 (3 min): Partially blend the chana dal with the mixture from the previous step.\newline
Step 2 (5 min): Slice the onions.\newline
Step 3 (3 min): Once the mixture is ready, you need to mix the mixture with cut onion.\newline
Step 4 (3 min): Later add some coriander and curry leaves and continue mixing.\newline
Step 5 (5 min): Heat up some oil in a pan.\newline
Step 6 (5 min): Shape the paste into circular disk-shaped chunks.\newline
Step 7 (5 min): Deep fry the shaped chunks in the hot oil.\newline
Step 8 (10 min): Fry the vada in oil until it turns golden brown.\newline
Step 9 (5 min): Serve the dish hot and along with some ketchup and some mint chutney.\newline\newline
Interruptible steps: 0, 1, 2, 3, 4, 6.\newline
Autonomous actions: step 5, 7, 8.\newline
Action Dependency: 0->1, 1->3, 2->3, 3->4, 4->6, 5->7, 6->7, 7->8, 8->9.\newline
Time Constraints: 5-7 (5 min), 7->8 (5 min), 8->9 (5 min).\newline
Steps 5, 7, 8 require stove.
} &{\#\# Recipe 2:Daikon-Radish\newline
Step 0 (5 min): Peel the skin from the radishes and rinse them.\newline
Step 1 (5 min): Slice them into thin circular slices.\newline
Step 2 (5 min): Chop napa cabbage into thin slices.\newline
Step 3 (5 min): Peel the skin off the onions and slice into cubes.\newline
Step 4 (3 min): Thinly slice the green onions.\newline
Step 5 (8 min): In a large non-stick pan, sauté the onion until slightly golden brown.\newline
Step 6 (2 min): Once the onion is cooked, place the sliced napa cabbage onto the pan.\newline
Step 7 (2 min): Stir fry for about 2 minutes.\newline
Step 8 (8 min): Add the radish into the pan and stir fry until soft.\newline
Step 9 (2 min): To season and garnish, add a couple of teaspoons of soy sauce, half a teaspoon of sesame oil, and a pinch of salt and pepper.\newline
Step 10 (3 min): Place the bacon slices onto a pan.\newline
Step 11 (10 min): Cook until crispy.\newline
Step 12 (5 min): Slice into smaller pieces to make bacon bits.\newline
Step 13 (2 min): Top the radish dish with some bacon bits.\newline\newline
Interruptible steps: 0, 1, 2, 3, 4, 12, 13.\newline
Autonomous actions: step 5, 11.\newline
Action Dependency: 0->1, 1->2, 2->3, 3->4, 4->5, 5->6, 6->7, 7->8, 8->9, 10->11, 11->12, 12->13, 9->13.\newline
Time Constraints: 5->6 (2 min), 6->7 (1 min), 7->8 (3 min), 8->9 (2 min).\newline
Steps 5, 6, 7, 8, 9, 11 require stove.
} \\
\midrule
\multicolumn{2}{c}{
    \includegraphics[width=0.9\textwidth]{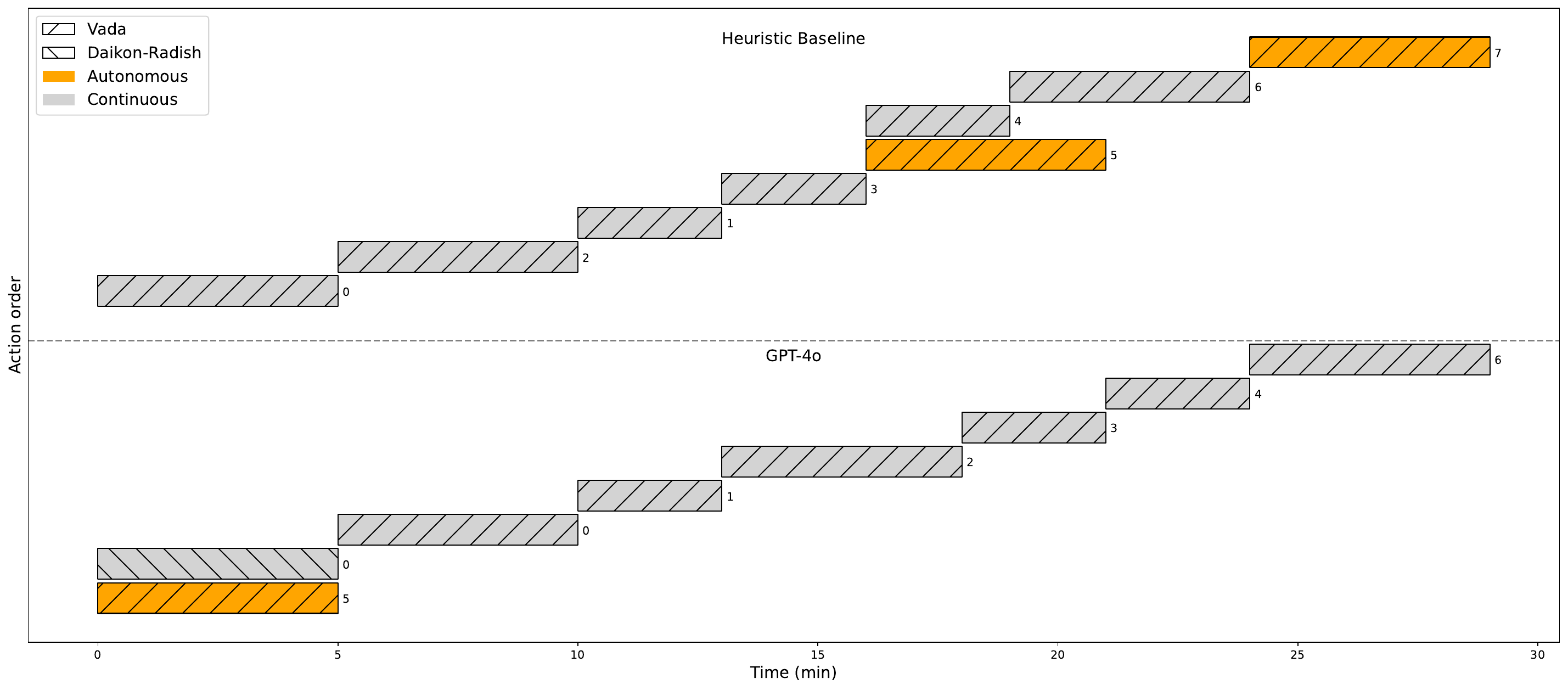}
}\\
\bottomrule 
\end{tabularx}
    \caption{Case study of planning with time constraints. We only show part of the plan to focus on the source of task failure. GPT-4o plans to execute step 5 of Vada first to maximize local efficiency which leads to violation of time constraints between step 5 and step 7 of Vada. The heuristic plan strategically starts step 5 of Vada until other prerequisite actions of step 7 are nearly finished.}
    \label{tab:case_w_tc} 
\end{table*}

\section{Details for Dataset Construction}
\subsection{Recipe Collection}
\label{subsec:recipe_collection}
We collect cooking recipes from existing benchmark MM-Res~\cite{pan2020multi}. MM-Res contains 9,850 recipes from cooking websites and has annotated the dependent relationship between actions in the recipe.
To curate cases from the MM-Res dataset for the purpose of our benchmark. We sample recipes that involve using a microwave, an oven or a stove and disregard those with more than 30 actions. We remove actions that is a non-cooking steps, such as introductory phrases like \textit{today we want orka}. Next, we ensure there are no temporal inconsistencies between steps. Optional statements are either removed or converted into mandatory steps. For example, \textit{You can use a spoon to get all the contents if needed} is revised to exclude \textit{if needed}. Actions are split for clarity if needed. for instance, \textit{boil water and pour water into a cup} becomes \textit{boil water} and \textit{pour water into a cup}, to separate the autonomous and continuous actions. Conversely, steps that describe sequential actions in separate sentences are merged; for example, \textit{use water to strain} and \textit{strain until the juices are gone} are combined into a single step. The dependent relationships of the revised actions are adjusted accordingly.

\subsection{Details for Recipe Annotation}

We recruit three graduate students with cooking experience to annotate the action properties and constraints in the recipes following the guidelines in Table~\ref{tab:annotation_guideline}. Each student identifies whether actions were autonomous or continuous, marks actions as interruptible or non-interruptible, and specifies any physical or time constraints associated with each action. To ensure consistency and accuracy, annotations were cross-verified among the annotators, with discrepancies resolved through discussion.


\subsection{Action Duration Distribution}
\begin{figure}
    \centering
    \includegraphics[width=1\linewidth]{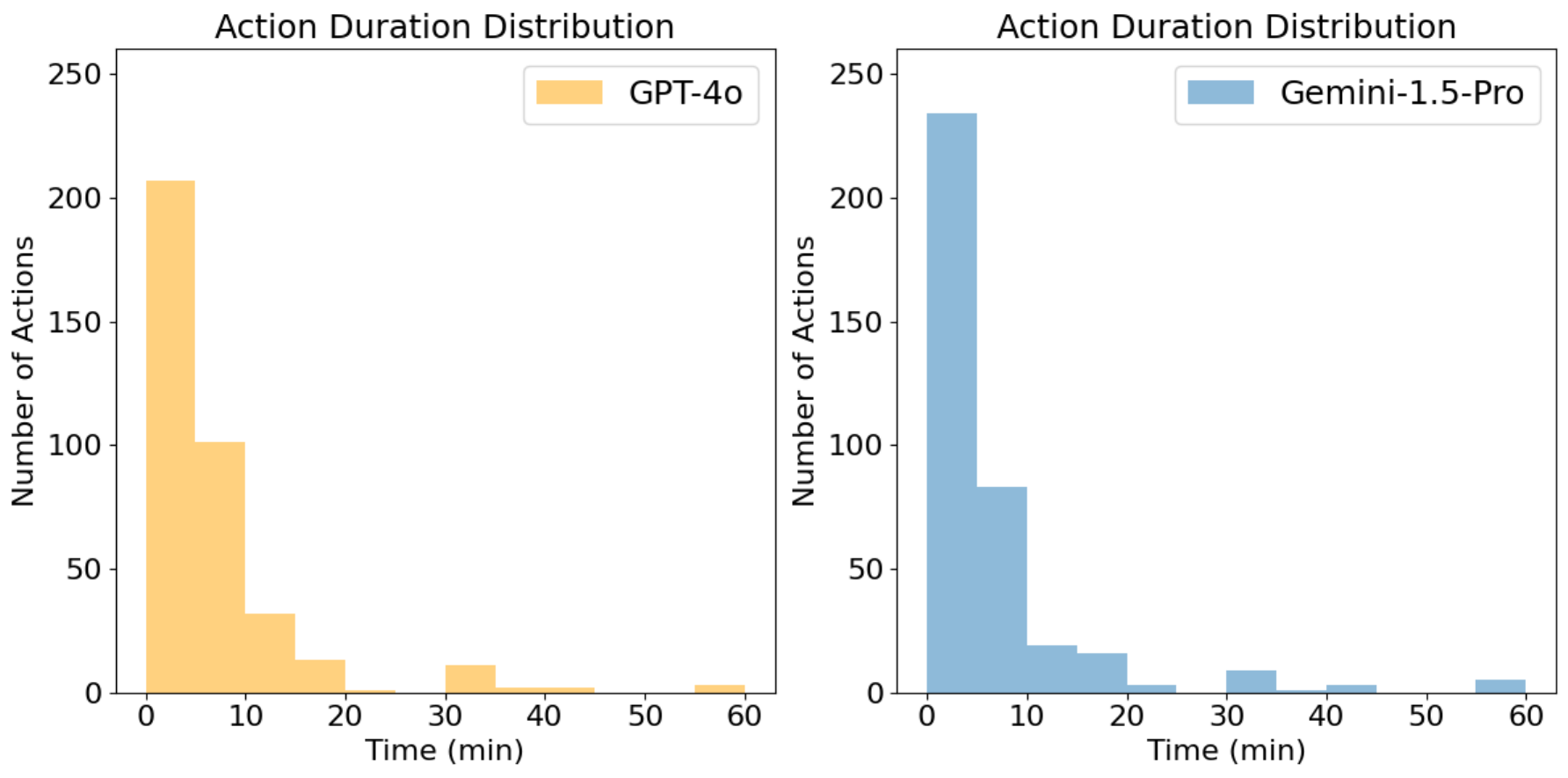}
    \caption{Distribution of action duration that Large Language Models annotate automatically. GPT-4o and Gemini-1.5-Pro share a similar distribution.}
    \label{fig:duration}
\end{figure}

We illustrate the distribution of action durations in Figure~\ref{fig:duration}, showing that the majority of actions take less than 10 minutes to execute. To investigate potential biases introduced by the use of GPT-4 in duration annotation, we also employed Gemini-1.5-Pro to annotate the durations. GPT-4 outperforms Gemini-1.5-Pro in both feasibility and efficiency (Table~\ref{tab:main}) despite both models exhibiting similar distributions in terms of action duration. It suggests that the automatic annotation does not introduce biases that could influence planning outcomes.

\section{Design of Metrics for Efficiency}
\label{appsec:efficiency}

Our metric is inspired by the Complete Speed (CS) metric from TimeArena~\cite{zhang2024timearena} but addresses unit and scale inconsistency issues in CS with the following calibration:\\
\noindent $\bullet$ \textbf{Unit Calibration}:  The progress score of each action in TimeArena is defined as:
\begin{equation}
    s=\frac{t_{i}}{\sum_{j=1}^{n} t_{j}} \times 100\%
\end{equation}
CS is then computed as the average of the highest (progress) score $P_i$ divided by the time taken to achieve it $T_i$ (min). TimeArena's CS divides a percentage by time, resulting in a unit of $\text{min}^{-1}$.
\begin{equation}
    CS = \frac{\sum_{i \in N} P_i}{\sum_{i \in N}T_{i}}
\end{equation}
To address this, we should replace the numerator with the action duration and ensure that the metric's unit is a percentage.

\noindent $\bullet$ \textbf{Scale Calibration:} CS can be biased by the duration of autonomous actions in the recipe, as more autonomous actions lead to a higher upper bound for efficiency.
To mitigate this limitation, the proportion of saved time $T_{save}$ and the cumulative duration of the executed autonomous actions $T_{auto}$ to scale the upper bound of efficiency to 1 for all recipe combinations.

\section{Implementation Details}
\subsection{Heuristic Baseline Algorithm}
We adapt the heuristic method from \citet{zhang2024timearena} to search for an efficient plan that is feasible. The details of the algorithm is presented in Algorithm~\ref{alg:heuristic}.
\label{appsub:heuristics}

\subsection{Environment}
\label{appsub:environment}
We implement an environment to provide feedback to the agent. The examples of feedback are listed in Table~\ref{tab:feedback_examples}. If the action fits all the constraints, the agent receives a message of the successful execution. And an observation of the action concurrency of the executed action, current timestamp, status of physical objects and the executing autonomous actions. If the action can not be executed, the environment will return an error message and detailed feedback about the violated constraint. We also provide a hint about executable actions to evaluate the global planning abilities of the agent solely in \S\ref{subsec:global_planning}. During the interaction with the agent, the maximum number of revisions is 10. Exceeding this number will be considered as task failure. And we abort the multitasking process if the agent attempts to execute the same action three times or violates any time constraints.

\subsection{Model Details}
\label{appsub:model_version}
We use the Instruct version for all sizes of Qwen2.5\footnote{\url{https://huggingface.co/Qwen}} and Llama-3.1\footnote{\url{https://huggingface.co/meta-llama}} models in our study. We use vllm~\cite{kwon2023efficient} to deploy Qwen2.5-7B, Llama-3.1-8B and Qwen-2.5-32B on a single A800 GPU, and Llama-3.1-70B and Qwen2.5-72B on four A800 GPUs.  We use the gpt-4o-2024-08-06 for GPT-4o\footnote{\url{https://platform.openai.com/docs/models\#gpt-4o}}, gpt-4o-mini-2024-07-18 for GPT-4o-mini\footnote{\url{https://platform.openai.com/docs/models\#gpt-4o-mini}}, Gemini-1.5-Pro-002 (2024-09-24) for Gemini-1.5-Pro\footnote{\url{https://ai.google.dev/gemini-api/docs/models/gemini\#gemini-1.5-pro}} .
\subsection{Planning Methods}
\label{appsub:method_details}
\noindent \textbf{Commonsense Reasoning} We prompt the model with the same guidelines in \S\ref{subsec:recipe_collection} and one example to generate the beliefs of action concurrency, action dependency and resource limitations. The temperature are set as 0 for all models. The max tokens for generation are set as 128. 

\noindent \textbf{Open-Loop Planning}
We evaluate the open-loop planning methods to determine if current LLMs can plan action sequences without interacting with the environment. Given the complexity of our task, we implement a \textit{Plan-and-Solve} baseline. In this approach, the model generates beliefs about unwritten properties and constraints through commonsense reasoning and creates a coarse-grained plan to perform actions simultaneously, aiming to reduce total execution time. Finally, the agent writes a fine-grained action sequence following one example for execution as shown in Table~\ref{tab:plan_solve_prompt}. The temperature is also set as 0 and the maximum generation tokens is 2048.


\noindent \textbf{Closed-Loop Planning with ReAct}
In this approach, we add the beliefs of unwritten constraints from commonsense reasoning to the recipe description. Then the agent performs one action at a time and predicts the next action based on interaction with the environment. The agent receives feedback after each interaction. If an action fails, detailed feedback is provided, prompting the model to reflect on its beliefs about unwritten properties and constraints and adapt its multitasking plan dynamically. The interaction continues until the agent believes all recipes are completed or the interaction is aborted by the environment. We set the temperature as 0 and the maximum generation tokens as 512. We parse the response to get the first action to avoid action trying to execute multiple actions during one interaction. The prompt for the react setting is detailed in Table~\ref{tab:react_prompt}. The prompt for ReAct with \textit{feasibility priority} setting is detailed in Table~\ref{tab:feasible_react_prompt}. In this setting, we prompt the model to finish recipes one by one to avoid violations of time constraints due to multitasking.

\section{Commonsense Reasoning Evaluation}
\label{appsec:commonsense}
We present the results of our evaluation for identifying unwritten properties and constraints in Table~\ref{tab:commonsense_result}. Most of the tested models display an F1 score above 80\% for identifying action dependency and object occupancy, with GPT-4o demonstrating robust performance by achieving F1 scores of 90.34\% and 91.12\%, respectively. Qwen2.5-32B and Qwen2.5-72B also exhibit strong commonsense capabilities in action dependency and resource limitations. However, the task of identifying autonomous actions poses a greater challenge. GPT-4o-mini achieves the highest recall at 82.18\%, while Gemini-1.5-Pro exhibits the highest precision at 88.33\%. While the models perform commendably in identifying action dependencies and object occupancy, there is a clear need for improvement in identifying autonomous actions, which present significant opportunities for multitasking.

\section{Open-Loop Planning Results}
\label{appsec:open_loop_results}
\noindent \textbf{LLMs cannot plan feasible multitasking without environmental feedback.} In the Plan-and-Solve setting, the model is prompted to plan an action schedule without feedback from the environment. The results in Table~\ref{tab:plan_solve_result} show that even GPT-4o achieves only a success rate of 3.1\% and a complete ratio of 21.7\% in the \textit{w/o time constraints} scenario. When time constraints are added, the success rate and complete ratio drop further. Many generated plans attempt to execute continuous actions simultaneously, leading to plan failure. This suggests that current LLMs lack the planning ability to schedule multitasking without environmental feedback.

\section{Planning Multitasking with Iterations}
The LLM-Modulo framework~\cite{gundawar2024robustplanningcompoundllm,kambhampati2024can} has demonstrated that large language models (LLMs) can effectively plan complex schedules with the help of critics, as seen in benchmarks like TravelPlanner~\cite{xietravelplanner}, which includes multiple soft and hard constraints. In our experiments described in \S\ref{sec3:experiments}, the agent cannot recover from time constraint violations. This section evaluates whether the LLM-Modulo framework can improve model performance by providing detailed critiques of the entire plan, thus enabling more robust global planning, correcting time constraint violations, and achieving efficient and feasible multitasking.

Rather than using a step-by-step evaluation as in \S\ref{sec3:experiments}, we design different critics to assess each property and constraint outlined in \S\ref{sec:property} and \S\ref{sec:constraints} for detailed feedback on the whole plan including the actions after the step that the plan fails. Following the approach in \citet{gundawar2024robust}, we use a reformatter to control the output format and concatenate the feedback from these critics and incorporate it into the initial prompt, performing 10 iterations. We evaluate GPT-4o and the latest o1-mini under this framework.

\noindent \textbf{Iterations of critics can help LLMs plan feasible multitasking.} We sample 20 cases from our benchmark which GPT-4o fails to solve in \textit{ReAct} setting. To evaluate the global planning abilities with iterations fairly, we prompt the models with the same plan and let the model revise the plan. We find that GPT-4o and o1-mini can deliver 3 feasible plans after 10 iterations. Out of the feasible plans, o1-mini achieves a relative efficiency of 98.89\% which is very close to a heuristic baseline.

\begin{table}[t]
   \centering
   \small
   \begin{tabular}{lcc}
   \toprule
   \textbf{Model} & \textbf{Pass Cases} & \textbf{R-Efficency$^{*}$} \\
   \midrule
   \multicolumn{3}{l}{\textit{Fix Failed Plans}} \\
   \midrule
   GPT-4o & 3 / 20 & 84.50\\
   o1-mini & 3 / 20& 98.89 \\
   \midrule
   \multicolumn{3}{l}{\textit{Optimize Feasible Plans}} \\
   \midrule
   GPT-4o & 10 / 18& 93.53 (\red{-2.62}) \\
   o1-mini & 8 / 18& 89.41 (\red{-3.87}) \\
   \bottomrule
   \end{tabular}
   \caption{Results of LLM-Modulo framework on generating feasible or more efficient plans. We report a number of feasible cases and the relative efficiency. $^{*}$: Average R-Efficiency for feasible plans.}. 
   \label{tab:llm_modulo}
   \end{table}

\noindent \textbf{LLMs can not adjust a feasible plan for higher multitasking efficiency without breaking feasibility} We prompt the model with 18 feasible plans from GPT-4o under ReAct setting, we then prompt the model to determine whether the plan can be further optimized for higher multitasking efficiency. GPT-4o and o1-mini can only maintain the feasibility of half of the plans and the relative efficiency of these also decreases compared with the initial plans.

\begin{table*}[t]
\centering
\small
\begin{tabular}{lcccccccccc}
\toprule
 \multirow{2}{*}{\textbf{Model}} & \multicolumn{4}{c}{\textbf{w/o time constraint}}  & \multicolumn{4}{c}{\textbf{w/ time constraint}} \\
\cmidrule(lr){2-5} \cmidrule(lr){6-9} 
& \multicolumn{1}{c}{\textbf{Success} }  & \multicolumn{1}{c}{\cellcolor{white}\textbf{Progress} }&   \multicolumn{1}{c}{\textbf{R-Efficiency} } & \multicolumn{1}{c}{\textbf{S$\times$E} } & \multicolumn{1}{c}{\textbf{Success}  }&  \multicolumn{1}{c}{\textbf{Progress}  }&  \multicolumn{1}{c}{\textbf{R-Efficiency}  } & \multicolumn{1}{c}{\textbf{S$\times$E} } \\
\midrule
& \multicolumn{8}{c}{\textit{Plan-and-Solve}}\\
\midrule
\quad Qwen2.5-7B &0.0 & 7.5 & 50.5 & 0.0 & 0.0 & 6.3 & 40.4 & 0.0 \\
\quad Llama-3.1-8B &0.0 & 6.2 & 38.6 & 0.0 & 0.0 & 6.6 & 42.9 & 0.0 \\
\quad Qwen2.5-32B & 0.0 & 8.6 & 78.1 & 0.0 & 0.0 & 8.8 & 54.6 & 0.0 \\
\quad Llama-3.1-70B &0.0 & 11.3 & 65.9 & 0.0 & 0.0 & 10.1 & 61.1 & 0.0 \\
\quad Qwen2.5-72B &1.5 & 17.2 & 87.5 & 1.5 & 0.0 & 13.3 & 76.6 & 0.0 \\
\textit{Closed-Source Models} & & & & & & & & \\
\quad GPT-4o-mini &1.5 & 13.4 & 78.0 & 1.4 & 0.0 & 10.3 & 63.6 & 0.0 \\
\quad Gemini-1.5-Pro & 0.0 & 6.1 & 61.4 & 0.0 & 0.0 & 6.6 & 68.3 & 0.0 \\
\quad GPT-4o &3.1 & 21.7 & 83.7 & 1.4 & 1.5 & 17.4 & 77.6 & 1.4 \\
\midrule
Heuristics  & 100 & 100 & 100 & 100 & 100 & 100 & 100 & 100 \\

\bottomrule
\end{tabular}
\caption{Results of Plan-and-Solve setting. We report Success Rate (\textbf{Success}), Average Progress Rate (\textbf{Progress}), Relative Multitask Efficiency (\textbf{R-Efficiency}) and Muititasking Ability (\textbf{S$\times$E}). }
\label{tab:plan_solve_result}
\end{table*}

\begin{table*}[t]
\centering
\small
\begin{tabular}{ccccc}
\toprule
  \textbf{Constraint} & \textbf{Model}  &  \textbf{Recall} &\textbf{ Precision} & \textbf{F1}\\
    \midrule
      \multirow{8}{*}{Action Concurrency}
& Qwen2.5-7B & 57.68 & 62.98 & 60.21 \\
& Llama-3.1-8B & 58.26 & 50.87 & 54.31 \\
& Qwen2.5-32B & 67.92 & 79.96 & 73.45 \\
& Llama-3.1-70B & 81.09 & 75.83 & 78.37 \\
& Qwen2.5-72B & 74.04 & 70.92 & 72.44 \\
& GPT-4o-mini & \textbf{82.18} & 68.27 & 74.58 \\
& Gemini-1.5-Pro & 40.48 & \textbf{88.33} & 55.51 \\
& GPT-4o & 73.10 & 79.39 & \textbf{76.12} \\
    \midrule
   \multirow{8}{*}{Action Dependency}
& Qwen2.5-7B & 69.67 & 78.15 & 73.67 \\
& Llama-3.1-8B & 79.68 & 82.61 & 81.12 \\
& Qwen2.5-32B & 87.41 & 92.18 & 89.73 \\
& Llama-3.1-70B & 91.57 & 92.58 & 92.07 \\
& Qwen2.5-72B &\textbf{ 92.18} & \textbf{92.82} & \textbf{92.50} \\
& GPT-4o-mini & 78.38 & 82.82 & 80.54 \\
& Gemini-1.5-Pro & 88.36 & 90.09 & 89.22 \\
& GPT-4o & 89.94 & 92.33 & 91.12 \\

   \midrule
      \multirow{8}{*}{Resource Limitations}
& Qwen2.5-7B & 69.82 & 91.26 & 79.12 \\
& Llama-3.1-8B & 85.31 & 88.96 & 87.10 \\
& Qwen2.5-32B & 91.28 & \textbf{96.15} & \textbf{93.65} \\
& Llama-3.1-70B & 88.30 & 97.21 & 92.54 \\
& Qwen2.5-72B & \textbf{92.22} & 95.04 & 93.61 \\
& GPT-4o-mini & 92.10 & 89.56 & 90.81 \\
& Gemini-1.5-Pro & 79.95 & 95.45 & 87.02 \\
& GPT-4o & 85.30 & 96.02 & 90.34 \\
\bottomrule
\end{tabular}
\caption{Results of commonsense reasoning for unwritten properties and constraints.}
\label{tab:commonsense_result}
\end{table*}

\begin{algorithm}[t]
    \caption{Heuristic Algorithm for Multitasking with time constraints}
    \label{alg:heuristic}
    \KwIn{Set of actions $\mathcal{A}$, Durations $\mathcal{T}$, Dependencies $p(\mathcal{A})$.}
    \KwOut{Heuristic minimal time $\mathcal{T}_{\text{min}}$.}
    \BlankLine
    Define autonomous actions $\mathcal{A}^\ast$ and continuous actions $\mathcal{A}'$ from $\mathcal{A}$.
    
    Sort $\mathcal{A}^\ast$ by $\mathcal{T}$ in descending order.
    
    $\mathcal{A} \leftarrow \text{concatenate}(\mathcal{A}^\ast, \mathcal{A}')$.
    
    Initialize \textit{Action\_list} as an empty list.
    
    \ForEach{$a_i \in \mathcal{A}$}{
        $P \leftarrow \text{BFS}(a_i, p(a_i))$ to collect prerequisites.
        
        \ForEach{$p_i \in P$}{
            \If{$p_i \in \mathcal{A}$}{
                \textit{Action\_list}.append($p_i$).
                
                Remove $p_i$ from $\mathcal{A}$.
            }
        }
        \textit{Action\_list}.append($a_i$).
    }
    
    Define function \textit{DFS}($\mathcal{A}^\ast$, $\mathcal{A}'$, $\mathcal{T}_{\text{min}}$):
    \If{\textit{Action\_list} is empty}{
        \Return $\mathcal{T}_{\text{min}}$.
    }
    
    \ForEach{$a_i \in$ \textit{Action\_list}}{
        \If{check\_constraint($a_i$)}{
            \If{$a_i \in \mathcal{A}^\ast$}{
                $\mathcal{A}^\ast \leftarrow \mathcal{A}^\ast \setminus \{a_i\}$.
                
                $\mathcal{T}_{\text{min}} \leftarrow \mathcal{T}_{\text{min}} $.
            }
            \Else{
                $\mathcal{A}' \leftarrow \mathcal{A}' \setminus \{a_i\}$.
                
                $\mathcal{T}_{\text{min}} \leftarrow \mathcal{T}_{\text{min}} + \mathcal{T}(a_i)$.
            }
            result $\leftarrow$ \textit{DFS}($\mathcal{A}^\ast$, $\mathcal{A}'$, $\mathcal{T}_{\text{min}}$)
            
            \If{result is not failure}{
                \Return result.
            }
        }
    }
    \Return failure.
    
    $\mathcal{T}_{\text{min}} \leftarrow 0$.
    
    result $\leftarrow$ \textit{DFS}($\mathcal{A}^\ast$, $\mathcal{A}'$, $\mathcal{T}_{\text{min}}$)
    
    \If{result is failure}{
        \Return "No feasible schedule found."
    }
    \Else{
        \Return result.
    }
\end{algorithm}

\begin{table*}
\centering
\small
\setlength{\tabcolsep}{5pt}
\begin{tabular}{p{2.0\columnwidth}r}
\toprule
\textbf{Action Concurrency}\\
Please identify if the action is autonomous or continuous.
\\ \quad - Autonomous Action: The action can be performed alongside other actions, allowing the agent to perform multiple tasks simultaneously. (e.g. preheat oven).
\\ \quad - Continuous Action: The step requires active involvement of the agent to complete and must be executed independently without overlapping with other tasks (e.g., 'Crack 3 eggs into a bowl').

\\ \\
\textbf{Execution Interruptibility}\\
A step classified as non-interruptible means that it cannot be split into two separate periods, and no other actions can be started during the execution of this action. Identify whether an action in a process can be interrupted or not
\\ \quad - If the action is logically interruptible (e.g., 'Dice the onions'), classify it as interruptible.
\\ \quad - If the action requires the agent to finish in one go(e.g., 'Keep stirring...'), classify it as non-interruptible.
\\ \quad - If the action involves heating (e.g., 'Melt the chocolate over low heat'), classify it as non-interruptible to ensure that the heating time is not extended.
\\ \quad - If the action can be executed in a short time (e.g., 'Pour water into a cup' or 'Add something into something'), classify it as non-interruptible.

\\\\
\textbf{Resource Limitations}\\
Annotate the steps that use one of the following physical objects.
\\ \quad - Oven: You should always preheat the oven to a specific temperature before using it. If the oven is already preheated by a previous step, you can skip the preheat action.
\\ \quad - Microwave: Use this tool to heat something quickly. You can only microwave for one recipe at the same time.
\\ \quad - Stove: Use the heater to warm your pan or pot for cooking.\\

\\ \\ 
\textbf{Time Constraints}\\
Identify pairs of actions if there is a time constraint between them. 
\\ \quad  - If the object of action has been heated, the time interval between steps should be some value to avoid extending the heating time (e.g., 'Fry the okra' -> 'Mix the onion with okra'). Steps involving cooling allow for more flexible time intervals.
\\ \quad - If the state of an object will change over time (e.g., 'Melt butter' -> 'Mix with something'), the next step should occur within a specific time frame to ensure the desired outcome
\\ \quad - Please only consider the actions with the direct dependent relationship. And you do not need to specify the time interval.\\
\bottomrule
\end{tabular}
\caption{Guidelines for recipe annotation.}
\label{tab:annotation_guideline}
\end{table*}

\begin{table*}
\centering
\small
\setlength{\tabcolsep}{5pt}
\begin{tabular}{p{2.0\columnwidth}r}
\toprule

\#\# Recipe 1:Baked-Potato\\
Step 0 (10 min): Preheat the oven to 425 degrees.\\
Step 1 (2 min): Pierce the potato several times with a fork.\\
Step 2 (5 min): Bake the potato in the preheated oven.\\
Step 3 (1 min): Melt butter in the microwave.\\
Step 4 (10 min): Remove potato from the oven and use a sharp knife to make decorative cuts on the top of the potato.\\
Step 5 (1 min): Pour melted butter over the potato and serve.\\
\\ 
- You can minimize the execution time based on the following properties:\\
You can execute only part of the action duration to pause steps 1, 4 for more efficient multitasking. But other actions must be finished without interruption.\\
 \\ 
- Do not violate any following constraints when executing this recipe:\\
Step 5 must be performed within 2 min after Step 3 is finished.\\ \\
Thoughts on the recipe: \\ 
The agent can perform autonomous actions step 0, 2, 3 in parallel with other actions to speed up the process.\\
The action before the arrow must be completed before the action after the arrow can be started: 0->2, 1->2, 2->4, 3->5, 4->5\\
The following actions would occupy the corresponding physical objects. The agent can not perform the action if the object is occupied. The properties such as volume and temperature of the object should also match the requirement of the recipe: Steps 0, 2 require oven, Steps 3 requires microwave\\
\bottomrule
\end{tabular}
\caption{Demonstration of recipe description. For the oracle setting, we replace the thoughts on the unwritten thoughts with the actual properties and constraints from the recipe}
\label{tab:description_example}
\end{table*}

\begin{table*}
\centering
\small
\setlength{\tabcolsep}{5pt}
\begin{tabular}{p{2.0\columnwidth}r}
\toprule

You are a multitask planner. You will plan an action sequence to finish some recipes as quickly as possible without violating any constraints.
\\ \\
\#\# Recipes\\
Each recipe is a sequence of actions designed to achieve a specific goal. Each action is a textual description companied with the duration to finish the action. Each recipe has autonomous actions such as boiling water that let the agent be idle during execution. They can be executed in parallel with other actions to speed up the process  Continuous actions such as pouring water occupy the agent and only one continuous action can be executed at the same time across all recipes.
\\ \\
\#\# Task Description\\
Your task is to complete all of the recipes as quick as possible while following the recipe. The key to success is to follow the recipe and constraints, then complete the steps in the correct order while minimizing the execution time by executing the autonomous actions concurrently.  First, let's analyze the recipe and create a concise plan on how to perform actions simultaneously to reduce the total execution time. Then write your action sequence following the plan. Your action should be a list of 'Step(step\_num, recipe\_name, time, timestamp)' which indicates the performing the given step for the given time at the timestamp. Your time and timestamp should be written as HH:MM:SS.
\\ \\
/*\\
\#\# Recipe 1:Baked-Potato\\
\textit{Description for baked potatoes}
\\\\
\#\# Recipe 2:Cheese-Sandwich\\
\textit{Description for cheese sandwiches}
\\ \\
\#\# Plan\\
Start by preheating the oven for both recipes simultaneously, since they require the same temperature and preheating is an autonomous action. This allows the agent to perform other actions while the oven is preheating.\\\\

While the oven is preheating, start the preparation steps that do not need the oven. I can pierce the potato (step 1 for Baked-Potato) and prepare the Cheese-Sandwich by buttering the bread (Step 1) and then placing the cheese between the slices (Step 2). These steps only take 5 minutes, then wait for the oven to preheat.\\\

At 00:10:00, the oven should be preheated. Bake the potato (Step 2 for Baked-Potato) and heat soup (Step 4 for Cheese-Sandwich) simultaneously. And wait for the autonomous actions to finish at 00:15:00 since the oven and microwave are occupied and no other actions can be performed.\\\\

At 00:15:00, start baking the sandwich (Step 3 for Cheese-Sandwich) and making decorative cuts on the potato (Step 4 for Baked-Potato) simultaneously. Step 4 of Baked-Potato is interruptible, so the agent can pause this action and perform step 5 of Cheese-Sandwich at 00:20:00 to serve the sandwich with the soup without violating the constraints. And the agent can continue to finish the potato with executing autonomous action step 3 and continuous action step 4 simultaneously and serve the potato at 00:26:00.\\\\

\#\# Action Sequence\\
Step(0, Baked-Potato, 10 min, 00:00:00), Step(0, Cheese-Sandwich, 10 min, 00:00:00), Step(1, Baked-Potato, 2 min, 00:00:00), Step(1, Cheese-Sandwich, 2 min, 00:02:00), Step(2, Cheese-Sandwich, 1 min, 00:04:00), Step(2, Baked-Potato, 5 min, 00:10:00), Step(4, Cheese-Sandwich, 5 min, 00:10:00), Step(3, Cheese-Sandwich, 5 min, 00:15:00), Step(4, Baked-Potato, 5 min, 00:15:00), Step(5, Cheese-Sandwich, 1 min, 00:20:00), Step(4, Baked-Potato, 4 min, 00:21:00), Step(3, Baked-Potato, 1 min, 00:25:00), Step(4, Baked-Potato, 1 min, 00:25:00), Step(5, Baked-Potato, 1 min, 00:26:00)\\
*/\\
Please follow the example to generate the action sequence for the following recipes.\\
/*\\
\#\# Recipe 1:\\
{Description}\\\\
\#\# Recipe 2:\\
{Description} \\\\
\#\# Plan\\
\bottomrule
\end{tabular}
\caption{Prompt for Plan-and-Solve setting.}
\label{tab:plan_solve_prompt}
\end{table*}

\begin{table*}
\centering
\small
\setlength{\tabcolsep}{5pt}
\begin{tabular}{p{2.0\columnwidth}r}
\toprule

You are a multitask planner. You will plan an action sequence to finish some recipes as quickly as possible without violating any constraints.
\\ \\
\#\# Recipes\\
Each recipe is a sequence of actions designed to achieve a specific goal. Each action is a textual description companied with the duration to finish the action. Each recipe has autonomous actions such as boiling water that let the agent be idle during execution. They can be executed in parallel with other actions to speed up the process  Continuous actions such as pouring water occupy the agent and only one continuous action can be executed at the same time across all recipes.
\\ \\
\#\# Task Description\\
You are required to analyze the current status of the environment and decide the next action to take so that you can finish the recipes in the shortest time without violating constraints. Give your thoughts on the given status, action history and observation. Your analysis should be within 100 words starting with 'Thought:'. Then you should choose your next action. If you think you have already completed all the recipes, please output 'Action: Finish'. Write your action as 'Action: Step(step\_num, recipe\_name, time, timestamp)' - Perform the given step for the given time at the timestamp. Your time and timestamp should be written as HH:MM:SS. You can only perform one action each time. Do not repeat actions that are already in progress. If you choose to wait for current actions to finish, please state the time you will wait for. Then write your next action as 'I will wait and perform the next action at HH:MM:SS. Action: Step(step\_num, recipe\_name, time, timestamp).
\\ \\
/*\\
\#\# Recipe 1:Baked-Potato\\
\textit{Description for baked potatos}
\\\\
\#\# Recipe 2:Cheese-Sandwich\\
\textit{Description for cheese sandwichs}
\\ \\

\#\# Action Sequence\\

...
\\
Observation: Continuous action executed successfully. The current timestamp is 0:5:0. Status of physical objects: oven is occupied, temperature is 425.0; microwave is not occupied; stove is not occupied; You are currently executing the following autonomous actions: step 0 of Baked-Potato (will finish at 0:10:0), step 0 of Cheese-Sandwich (will finish at 0:10:0)\\
Thought: I have completed the preparatory steps for the Cheese-Sandwich. Since both recipes are still preheating in the oven, I should wait for the preheating to complete. After preheating, both recipes are ready for the next baking step. Step 4 of the Baked-Potato is a long continuous step that can be executed in parallel with steps 3 and 4 of the Cheese-Sandwich. So I should bake the Baked-Potato first.\\
I will wait and perform the next action at 00:10:00. Action: Step(2, Baked-Potato, 00:05:00, 00:10:00) \\

...
\\
Observation: Autonomous action started successfully. The current timestamp is 0:15:0. Status of physical objects: oven is occupied, temperature is 425.0; microwave is not occupied; stove is not occupied; You are currently executing the following autonomous actions: step 4 of Cheese-Sandwich (will finish at 0:20:0)\\
Thought: With the Cheese-Sandwich currently baking in the oven, I can now focus on the the Baked-Potato. I will remove the Baked-Potato from the oven and prepare it for serving. I should not start melting the butter for the Baked-Potato before nearly finishing the Step 4 of Baked-Potato to avoid the butter cooling down. So I will start making decorative cuts on the top of the potato first. And I will stop at 0:20:00 to serve the Cheese-Sandwich to avoid violating the time constraint.\\
Action: Step(4, Baked-Potato, 00:05:00, 00:15:0)\\
...
\\

Observation: Continuous action executed successfully. The current timestamp is 0:21:0. Status of physical objects: oven is occupied, temperature is 425.0; microwave is not occupied; stove is not occupied\\
Thought: I have served the Cheese-Sandwich and can now continue with the Baked-Potato. I will prepare the melted butter and pour it over the potato to complete the recipe. Since the step 5 must be performed within 2 min after Step 3, I will start melting the butter when step 4 is nearly finished in 1 minute. Now I would continue with the decorative cuts on the top of the potato.\\
Action: Step(4, Baked-Potato, 00:04:00, 00:21:0)\\

...\\

Observation: Continuous action executed successfully. The current timestamp is 0:27:0. Status of physical objects: oven is occupied, temperature is 425.0; microwave is occupied; stove is not occupied\\
Action: Finish\\
*/\\
Please follow the example to generate the action sequence for the following recipes.\\
/*\\
\#\# Recipe 1:\\
\textit{Description}\\\\
\#\# Recipe 2:\\
\textit{Description} \\\\
\#\# Action Sequence\\
\bottomrule
\end{tabular}
\caption{Prompt for ReAct-style action sequence generation.}
\label{tab:react_prompt}
\end{table*}

\begin{table*}
\centering
\small
\setlength{\tabcolsep}{5pt}
\begin{tabular}{p{2.0\columnwidth}r}
\toprule
You are a multitask planner. You will plan an action sequence to finish some recipes without violating any constraints.
\\\\
\#\# Recipes\\
Each recipe is a sequence of actions designed to achieve a specific goal. Each action is a textual description companied with the duration to fininsh the action. Each recipe has autonomous actions such as boiling water that lets the agent be idle during execution. They can be executed in parallel with other actions to speed up the process  Continuous actions such as pouring water occupy the agent and only one continuous action can be executed at the same time across all recipes.\\\\

\#\# Task Description\\
You are required to analyze the current status of the environment and decide the next action to take so that you can finish the recipes without violating constraints. Do not rush to complete the recipes. Take your time to execute the actions without violating constraints. especially the time constraints. You should prioritize on not violating the time constraints and ensure the recipe is completed successfully.\\\\

Give your thoughts on the given status, action history and observation. If you find your initial thoughts of the recipes does not align with the current status, you can revise your initial thoughts. Your analysis should starts with 'Thought:'. Then you should choose your next action. If you think you have already completed all the recipes, please output 'Action: Finish'. Write your action as 'Action: Step(step\_num, recipe\_name, time, timestamp)' - Perform the given step for the given time at timestamp. Your time and timestamp should be written as HH:MM:SS. You can only perform one action each time. Do not repeat actions that are already in progress. If you choose to wait for current actions to finish, please state the time you will wait for. Then write your next action as 'I will wait and perform the next action at HH:MM:SS. Action: Step(step\_num, recipe\_name, time, timestamp).\\\\

/*\\
\#\# Recipe 1:Baked-Potato\\
\textit{Description for baked potatos}

Observation: The current timestamp is 00:00:00. Status of physical objects: oven is not occupied, is not preheated; microwave is not occupied; stove is not occupied\\
Thought: I will preheat the oven. This will allow me to perform other actions while the oven is preheating.\\
Action: Step(0, Baked-Potato, 00:10:00, 00:00:00)\\\\

Observation: The current timestamp is 00:00:00. Status of physical objects: oven is occupied, temperature is 425.0; microwave is not occupied; stove is not occupied; You are currently executing the following autonomous actions: step 0 of Baked-Potato (will finish at 00:10:00)\\
Thought: Since the oven is preheating, I can start piercing the potato for the Baked-Potato recipe. And this does not risk violating any time constraints.\\
Action: Step(1, Baked-Potato, 00:02:00, 00:02:00)\\\\

Observation: The current timestamp is 00:04:00. Status of physical objects: oven is occupied, temperature is 425.0; microwave is not occupied; stove is not occupied; You are currently executing the following autonomous actions: step 0 of Baked-Potato (will finish at 00:10:00), step 0 of Cheese-Sandwich (will finish at 00:10:00)\\
Thought: Step 1 of the Baked-Potato is complete. I will now move on to step 2 of the Baked-Potato. This does not risk violating any time constraints.\\
Action: Step(2, Baked-Potato, 00:05:00, 00:10:00)\\\\

Observation: The current timestamp is 00:10:00. Status of physical objects: oven is occupied, temperature is 425.0; microwave is not occupied; stove is not occupied; You are currently executing the following autonomous actions: step 2 of Baked-Potato (will finish at 00:15:00)\\
Thought: Step 2 of the Baked-Potato is complete. Since Step 5 depends on both Step 3 and Step 4 and Step 5 must be performed within 2 minutes after Step 3 is complete, and. Executing Step 3 first will result in violation of time constraints since Step 4 is a long continuous step. I will now first finish Step 4 of the Baked-Potato to ensure the time constraints are followed.\\
Action: Step(4, Baked-Potato, 00:10:00, 00:15:00)\\\\

...\\\\

Observation: The current timestamp is 00:26:00. Status of physical objects: oven is not occupied, temperature is 425.0; microwave is not occupied; stove is not occupied\\
Thought: The recipe is completed.\\
Action: Finish\\\\
*/\\
Please follow the example to generate the action sequence for the following recipes.\\
/*\\
\#\# Recipe 1:\\
\textit{Description}\\\\
\#\# Action Sequence\\
\bottomrule
\end{tabular}
\caption{Prompt for ReAct-style action sequence generation with \textit{feasibility priority}.}
\label{tab:feasible_react_prompt}
\end{table*}

\begin{table*}
\centering
\small
\setlength{\tabcolsep}{5pt}
\begin{tabular}{p{0.6\columnwidth}<{\centering}p{1.5\columnwidth}}
\toprule
\multicolumn{2}{c}{\textbf{Feedback}}\\
\midrule
Wrong Recipe & Recipe C is not one of our goals. Please select actions from our recipes A and B \\
Wrong Action & There is no step X in recipe A.\\
Time Error & The current time is HH:MM:SS. You can not perform any actions before the current timestamp. \\
Infeasible Multitask & You can not perform step X of Recipe A and step Y of Recipe B simultaneously since they are all continuous actions.\\
Object Occupancy & Object M is currently occupied.\\
Object Temperature & The Property of the Object is $\alpha$, but step X of recipe A needs $\beta$. \\
Object Volume & The object only has a volume of water. It is not enough for this action. \\
Dependency & Step Y of recipe A can not be performed because prerequisite step X is not completed. (The expected finish time is HH:MM:SS) \\
Repeated Action & Prerequisite step X is already used for the next action step Y in recipe A. You should not execute the same step twice. If you insist, please complete all the previous steps first.\\
time constraint & The time interval between Step X and Step Y in Recipe A exceeds the allowed time limit t min.\\
Action Duration & Your plan execution time t min exceeds the time needed to perform the action. \\
Execution Interruptibility & Step X of Recipe A is not interruptable. You should finish the action in one go.\\
\midrule
\multicolumn{2}{c}{\textbf{Observation}}\\
\midrule
Success Execution & Autonomous / Continuous action executed successfully. Stove is not occupied; Oven is not occupied, temperature is $t$. You are executing step X of recipe A.\\
Failed Execution & Step X of Action A can not be executed. \\
\midrule
\multicolumn{2}{c}{\textbf{Hint}}\\
\midrule
Executable Actions & The following actions are ready to be executed after HH:MM:SS, Step X of Recipe A, Step Y of Recipe B.\\
\bottomrule
\end{tabular}
\caption{Examples of observation, feedback and hints from the environment.}
\label{tab:feedback_examples}
\end{table*}

\begin{table*}
\centering
\small
\setlength{\tabcolsep}{5pt}
\begin{tabular}{p{2\columnwidth}}
\toprule
\textbf{Critic Example}\\
\midrule
\textbf{\#\# Critic for the plan}:\\ \\
\textbf{Plan Completeness}: The following actions are missing in your plan: step 2 of Cobbler; step 3 of Cobbler; step 4 of Cobbler; step 5 of Cobbler; step 6 of Pancakes; step 7 of Pancakes. Include them in the plan to complete the recipe.\\ \\
\textbf{Action Duration}: The duration of the following actions do not align with the action duration: Pancakes step 9; Pancakes step 10; Pancakes step 14; Pancakes step 12. Make sure the duration of the actions are correct.\\\\
\textbf{Action Concurrency}: You can not start another action while executing a continuous action. In your plan, the following actions can not be performed simultaneously with each other: step 0 of Pancakes and step 1 of Pancakes; step 1 of Pancakes and step 5 of Pancakes ; step 6 of Cobbler and step 3 of Cobbler . Please adjust the timeline to avoid the conflict. \\ \\
\textbf{Action Interruption}: The following actions should not be interrupted in your plan: step 14 of Pancakes. Make sure they are finished in one go.\\ \\
\textbf{Action Dependency}: step 1 of Pancakes can be performed only after prerequisite action step 0 of Pancakes is finished. You should complete the prerequisites before performing the next action.\\\\
\textbf{Time constraint}: The following action pairs violate the time constraint: step 5 of Cobbler should start within 2 min after step 3 of Cobbler is finished; step 14 of Pancakes should start within 2 min after step 12 of Pancakes is finished. Reschedule the actions to meet the time constraint.\\\\
\textbf{Physical Object}: Step 14 of Pancakes can not be performed at time 00:02:00 due to Object stove is occupied. Adjust the use of the physical objects to meet the requirements.\\
\\
\textbf{Multitasking Efficiency} The plan is feasible. The agent is idle during the following timestamps: HH:MM:SS and HH:MM:SS. You can assign continuous actions to the agent to optimize the plan for a shorter execution time. If you think the plan is optimal, you can answer Action: Done to finish the task.\\
\bottomrule
\end{tabular}
\caption{Critic example for LLM-Modulo framework.}
\label{tab:critic_example}
\end{table*}

\end{document}